\documentclass[11pt]{article}
\usepackage[utf8]{inputenc}
\usepackage{graphicx}
\usepackage{hyperref}
\usepackage{amsmath, amssymb}
\usepackage{geometry}
\usepackage{datetime}
\usepackage{subcaption}
\usepackage[style=numeric, sorting=none, maxbibnames=3, minbibnames=1, backend=bibtex]{biblatex}
\usepackage{authblk}

\newcommand{\betavae}{$\beta$-VAE}

\title{Spectral Considerations on Unsupervised Time-domain EEG Representation Learning with Deep Networks}

\author[1,2]{David Aquilué-Llorens}
\author[1]{Abel Puigseslloses}
\author[1]{Angel Blanco}
\author[1]{Claire Braboszcz}
\author[1,*]{Aureli Soria-Frisch}

\affil[1]{Starlab Barcelona S.L., Neuroscience Business Unit, Barcelona, Spain}
\affil[2]{Center for Brain and Cognition, Computational Neuroscience Group, Universitat Pompeu Fabra, Barcelona, Spain}
\affil[*]{Corresponding Author: aureli.soria-frisch@starlab.es}

\date{}

\begin{document}

    \maketitle
%     \section*{Modifications in Last Version}
%     \begin{itemize}
%         \item Added small intro in paragraph discussing inclusion of REVE Foundation Model.
%         \item Added REVE short description in Methods.
%         \item Moved Topographic PSD attenuation (Methods Section and Previous Figure 5) to Appendix, now showing only PSDs which need no methods.
%         \item Results 3.1 Added REVE results (text, Figure 1, Figure 2).
%         \item Results 3.2 Added reference Appendix on Spectral Bias in REVE (results of new REVE spectral bias without showing bias plus the masked spectral bias for the autoencoders).
%         \item Results 3.3 Added REVE UMAP BBHI Results (text, Figure 4) + Added Figure 5 of PSD attenuation for the three autoencoder models on the BBHI dataset.
%         \item Adapted Discussion
%         \item Adapted Conclusions
%         \item Changed what we previously named "Spectral Bias" to "Spectral Learning Order"
%     \end{itemize}

% \clearpage

\paragraph*{Abstract}
Electroencephalography (EEG) is a non-invasive, scalable source of quantitative biomarkers for the early diagnosis of brain disorders, and deep representation learning has become a powerful paradigm for extracting information-rich latent features from raw signals. However, neural networks tend to under-represent high-frequency information, a concern that is especially acute for EEG given its oscillatory nature and characteristic $1/f^\chi$ power decay, which risks masking disease-relevant high-frequency signatures. In this study, we systematically investigate the spectral constraints of unsupervised time-domain feature extraction across three autoencoder architectures (LSTMVAE, VAEEG, Transformer) and a state-of-the-art EEG foundation model. Using synthetic experiments and empirical validation, we show that while autoencoders capture macro-scale shifts such as alpha slowing, they remain biased against higher frequencies due to a magnitude-driven masking effect induced by the Mean Squared Error loss under 1/f-decaying spectra. Increasing latent dimensionality mitigates this attenuation but sacrifices latent compactness and generative stability. We argue that, as the field advances toward large-scale foundation-level feature extractors, it is imperative to move beyond simple time-domain reconstruction and to integrate spectral-domain loss terms and gradient-balancing techniques, so that the subtle high-frequency transients critical for diagnosing complex brain disorders are preserved in the latent representation.

\section{Introduction}

    Brain disorders, including neurological and mental health
    conditions, fundamentally disrupt healthy neural function,
    negatively impacting the quality of life of both
    patients~\cite{Momen2020, Vollset2024, Xing2024} and their
    caregivers~\cite{Krmer2024}.
    Collectively, these conditions now rank among the top ten leading causes of
    disability worldwide~\cite{Harris2024} and suppose an enormous economic burden, e.g.  469.5 million disability-adjusted life-years (DALYs) globally, total cost estimates reaching 1.7 trillion EUR in Europe \cite{Li2025}. These facts  highlight the urgent clinical
    need for more effective diagnosis and intervention.

    Despite this urgency, current diagnostic procedures remain significantly limited.
    Standard procedures often rely on techniques that are either invasive,
    such as brain tumor biopsies or cerebrospinal fluid extraction for
    neurodegenerative biomarkers~\cite{Jack2024}; expensive and resource-intensive,
    such as Positron Emission Tomography (PET) 
    for Alzheimer’s Disease~\cite{Johnson2013, Wittenberg2019}; or inherently time intensive and prone
    to subjectivity, as seen in traditional neuropsychology questionnaires and
    cognitive evaluations~\cite{Tombaugh1992, Insel2010, NieuwenhuisMark2010}.

    Consequently, there is an ever-increasing need for robust, quantitative, and
    objective biomarkers capable of facilitating precise diagnosis in the
    preclinical or early stages of disease, the therapeutic window where disease
    modifying therapies are most effective in mitigating symptoms and delaying
    progression~\cite{Grill2021, Cummings2024, Clarke2025}.\\

    In this landscape, Electroencephalography (EEG) emerges as a uniquely powerful
    candidate.
    As a direct measure of brain activity, EEG offers a non-invasive, scalable, and
    cost-effective solution that requires minimal patient cooperation.
    Its high temporal resolution and portability enable continuous, remote, and at-home
    monitoring~\cite{Casson2019, Sohrabpour2025}.

    While EEG has long been utilized in clinical settings for the management of
    epilepsy and sleep disorders~\cite{berry2012aasm, Tatum2021}, translational
    efforts to leverage its full diagnostic potential for other pathologies are still
    ongoing.
    To bridge the gap between research and routine clinical practice, we must
    discover novel, generalizable biomarkers that can reliably stratify complex
    brain-related diseases, show enough sensitivity-specificity for diagnosis procedures, dynamically capture the effect of therapeutical interventions, contextualize mechanistic hypotheses of pathophysiology, and add explainable interpretations of clinical observations at brain activity level.\\

    Traditionally, biomarker discovery has relied on feature extraction
    guided by specific domain knowledge.
    However, this paradigm is rapidly being surpassed by automated Artificial Intelligence
    approaches~\cite{Roy2019, Hosseini2021}.
    Propelled by the exponential rise of Deep Learning and the increasing
    availability of large-scale, open-source EEG
    repositories~\cite{Obeid2016, Babayan2019, vanDijk2022, Zhang2025},
    data-driven methods are now capable of uncovering complex novel patterns~\cite{Chen2025}.

    The Deep Learning field has provided a wide array of tools capable of
    distilling relevant information from noisy, high-dimensional natural signals
    (text, image, audio, physiology), through the process of ``encoding``.
    Representation Learning approaches are capable of compressing raw inputs into compact,
    information-rich latent spaces~\cite{Bengio2012}.

    In particular, Variational Autoencoders (VAEs) have proven particularly
    successful for unsupervised feature extraction.
    By training to encode (compress) and subsequently decode (reconstruct) input
    data, VAEs learn to approximate the underlying data distribution.
    $\beta$-Variational Autoencoders further enhance this capability by
    disentangling the generative factors of the data, improving the interpretability
    of the latent space as well as the fidelity of the model's generative
    output~\cite{higgins2017betavae, Doersch2016}.

    Beyond feature extraction, the generative nature of VAEs allows the
    synthesis of novel data points via latent space
    sampling~\cite{Kingma2019, Huang2025}.
    This capability paves the way for advanced data augmentation strategies,
    essential for training downstream classifiers or increasing diversity in large
    datasets used to train emerging foundation models~\cite{Kang2025}.\\

    However, current research has brought to light an artifact of VAEs: a tendency
    to lose fine-grained detail and high-frequency information, resulting in
    ``smoothed`` reconstructions that capture only the coarse characteristics of the
    input~\cite{Bjork2022, Xiang2025, Lai2025}.

    An additional, overlapping issue, more general to neural network training, called
    ``Spectral Bias`` has shown that networks trained through gradient descent and
    backpropagation prioritize learning low frequency components early in
    training~\cite{Rahaman2018, Xu2019}.

    For EEG analysis, these effects pose a significant challenge.
    EEG is inherently oscillatory, with critical pathological information
    distributed across frequency bands ranging from 0.5 to 45 Hz.
    Distinct disorders manifest as spectral shifts in specific ranges.
    For instance, Alzheimer's Disease has been widely reported to show a slowing of the
    alpha oscillations~\cite{Kopanov2024, Sohrabpour2025} while Major
    Depressive Disorder is often associated with alterations in theta and gamma
    power~\cite{deAguiarNeto2019}.
    Crucially, the EEG power spectrum follows a $1/f^{\chi}$ decay, meaning
    the signal power diminishes rapidly as frequency increases.
    This natural decay risks exacerbating the frequency-specific limitations of deep
    learning models, potentially masking high-frequency biomarkers vital for
    brain disorder diagnosis.\\

    In this study, we systematically evaluate the capacity of deep autoencoders to
    distill high-dimensional EEG signals into information-rich latent spaces,
    specifically addressing whether temporal-domain reconstruction can resolve the
    nuanced spectral signatures required for clinical stratification.
    Evaluating three different autoencoder architectures (LSTMVAE, VAEEG,
    Transformer), we investigate for the first time the ``frequency-specific biases`` inherent in neural network
    training and how they interact with the characteristic $1/f$ power decay of human electrophysiology. 

    We additionally extend our analyses to a state-of-the-art EEG foundation model~\cite{elouahidi2025reve}. Trained on large unlabelled corpora with self-supervised objectives, foundation models have become the new standard across a wide range of neuroimaging tasks, including sleep-staging, brain-computer interfaces, epilepsy monitoring, and affective assessment~\cite{Chen2025, elouahidi2025reve}. Because they are trained as universal feature embedders rather than for specific tasks, we ask whether foundation models suffer from spectral biases similar to those in application-specific autoencoders.

    Our work dissects theoretical challenges to be addressed by modern Representation Learning approaches to become effective hypothesis generating machines with clinical and diagnostic value.

    Through a series of synthetic experiments simulating a cognitive decline
    spectrum plus validation on an empirical dataset \cite{Cattaneo2018, PerellnAlfonso2024}, we demonstrate that while reconstruction autoencoders effectively capture macro-scale spectral characteristics, they suffer
    from significant magnitude-driven masking.
    Our findings reveal that the standard Mean Squared Error (MSE) loss prioritizes
    high-power, low-frequency components, effectively ``smoothing`` out the subtle,
    high-frequency transients essential for diagnosing complex brain disorders.
    By quantifying the trade-offs between latent dimensionality and spectral
    fidelity, this work identifies critical limitations in current unsupervised
    feature extraction methods and proposes a roadmap for more robust,
    frequency-aware EEG foundation models.

\section{Methods}\label{sec:methods}

\subsection{Synthetic EEG-like data generation}\label{subsec:synth-eeg-alpha}
    To evaluate the feature distillation capabilities of the proposed autoencoders
    under controlled conditions, we generated a synthetic dataset modeling the
    electrophysiological characteristics of a dementia cohort.
    This dataset simulates four stages of cognitive decline: Healthy Control (HC),
    Subjective Cognitive Decline (SCD), Mild Cognitive Impairment (MCI), and
    dementia by Alzheimer’s Disease (AD).

    \paragraph{$\mathbf{\alpha}$ Slowing Dataset}
    The primary feature modeled is the progressive ``alpha slowing`` characteristic of
    neurodegeneration.
    For each synthetic subject $n$ in group $G$, a peak alpha frequency $f_\alpha^n$
    was sampled from a uniform distribution
    $\mathcal{U}(\mu_G - 0.25, \mu_G+0.25)$.

    The mean frequencies were set to
    $ \mu_{HC} = 12$ Hz, $\mu_{SCD} = 11$ Hz, $mu_{MCI} = 9.5$ Hz, $\mu_{AD} = 8 $ Hz
    to reflect the shift from typical alpha towards slower frequencies\cite{Sohrabpour2025}.

    For each subject we generated 500 epochs ($T=2$s, $f_s = 250$ Hz) across 19
    virtual channels.
    The signals were synthesized in the frequency domain by defining a Power
    Spectral Density (PSD) consisting of two elements:
    \begin{itemize}
        \item A $1/f^\chi$ ``pink noise`` background, where $\chi=0.75$
        \item A Gaussian-shaped peak, with $\sigma^2_\alpha = 1$, centered at $f_\alpha^n$ representing the alpha
        rhythm.
    \end{itemize}

    To ensure variability, a random phase $\phi \in [0, 2\pi]$ was independently
    assigned to each frequency bin and channel.
    The final time-series signals were obtained via the Inverse Fast Fourier
    Transform (IFFT).

    While this model simplifies the non-linear dynamics of alpha slowing and
    omits inter-channel correlations typical of empirical EEG \cite{Sohrabpour2025}, it provides a
    controlled ground-truth frequency shift to validate the autoencoder's latent space
    organization.

    \paragraph{$\mathbf{\beta / \gamma}$ Slowing Dataset}
    We generated a secondary dataset, following a near identical methodology to the
    \textit{$\alpha$ Slowing Dataset}, that simulates the slowing of a high-$\beta$/
    low-$\gamma$ rhythm, while maintaining a fixed alpha oscillation across all groups.
    The synthesized PSD for this dataset is composed of three additive elements:
    \begin{itemize}
        \item A $1/f^\chi$ ``pink noise`` background, where $\chi=0.75$
        \item A Gaussian-shaped peak, with $\sigma^2_\alpha=1$, centered at $f_\alpha = 10$ Hz representing a
        stable
        alpha rhythm, common to all subjects and groups.
        \item A Gaussian-shaped peak, with $\sigma^2_{\beta/\gamma}=1$, centered at $f_{\beta / \gamma}^n$
    \end{itemize}
    where now the subject-specific $f_{\beta / \gamma}^n$ is sampled for each
    from the group-specific uniform distributions, $\mathcal{U}(\mu_G - 0.25, \mu_G+0.25)$, with
    $\mu_{HC} = 32$ Hz, $\mu_{SCD} = 31$ Hz, $\mu_{MCI} = 29.5$ Hz, $\mu_{AD} = 28$ Hz.

    We incur here in an additional simplification where both the alpha and $\beta/\gamma$
    peaks in the PSD are of the same magnitude.
    Typically, amplitudes of high frequency EEG oscillations are significantly
    smaller than those of lower frequencies, which in turn adds increased difficulty
    to the feature extraction process.\\

    To simulate training/validation/test partitions we generated a training
    partition with 40 randomly generated synthetic subjects per group, a validation
    partition with 10 synthetic subjects per group, and a test partition with 10
    synthetic subjects per group.

    We provide in Figure~\ref{fig:synth_data_perf} the group-averaged PSDs for the
    $\alpha$ and $\beta / \gamma$ datasets, as well as, an example epoch for each dataset to visualize the qualitative differences in spectral content.

\subsection{BBHI Empirical EEG Dataset}
    The Barcelona Brain Health Initiative (BBHI) is a longitudinal study focused on identifying determinants of brain health. The BBHI cohort has included >4500 healthy participants aged 40–65 years followed through online questionnaires. We analyzed the EEG data of N=963 individuals from the Phase II study, where detailed in-person evaluations (including medical, neurological and psychiatric examinations, assessment of physical fitness, neuropsychological assessments, structural and functional brain magnetic resonance imaging, electroencephalography and perturbation-based non-invasive brain stimulation evaluations) were repeatedly conducted at two-yearly intervals \cite{Cattaneo2018}.  The EEG data was acquired from 32-channels placed on the scalp following the standard 10-20 system through the Enobio-32 device from Neuroelectrics (Barcelona, Spain) during resting state (eyes open and eyes closed) and during the performance of an auditory Oddball task and a Flanker task. Our study is based on the eyes closed EEG resting state data from the 19 standard channels in the 10-20 system.

\subsection{Autoencoder Models}\label{sec:autoencoders}

    An Autoencoder is a deep neural network, typically characterized by a symmetric
    structure, mirroring at the latent space layer between the two main subnetworks: the
    encoder and the decoder.
    The encoder aims to extract relevant information from input samples and encode it
    into the usually lower-dimensional latent space.
    The decoder uses the encoded embeddings and aims to generate an accurate
    reconstruction of the input sample.
    During training, network weights are updated in order to minimize the
    reconstruction error (typically the Mean Square Error, MSE) between the input
    and output (reconstructed) samples.

    The $\beta$ Variational Autoencoder (\betavae) is a modified probabilistic autoencoder where the
    LS embedding becomes an $N_{LS}$-dimensional probability distribution
    $\mathcal{N}(\mu, \sigma)$ with $\mu, \sigma$ determined by the encoder.
    A regularization is applied through the ELBO loss where the distance (
    KL-divergence) between the probabilistic embedding is minimized to a Gaussian
    distribution of null mean and identity standard deviation:

    \begin{equation}\label{eq:ELBO}
        \mathcal{L}_\beta = MSE(x, \hat{x}) + \beta KL(\mathcal{N}(\mu, \sigma) ||
        \mathcal{N}(\mathbf{0}, \mathbf{1})).
    \end{equation}

    The regularization applied through the KL-divergence homogenizes the Latent Space,
    demonstrably improving the feature extraction process as well as providing
    generative capabilities for the model.

    \subsubsection{LSTM - Variational Autoencoder}

    The LSTM-based variational autoencoder (LSTMVAE) presented in this study builds
    upon our previous work on autoencoder architectures for EEG
    denoising~\cite{Aquilue2025}.
    The network takes as input a multi-channel EEG epoch
    $\in \mathcal{R}^{N_C \times N_T}$, where $N_C$ denotes the number of channels
    and $N_T$ the number of time samples, and maps it into a low-dimensional latent
    space through an encoder–decoder framework.

    The encoder consists of two concatenated LSTM layers, with hidden state sizes
    $N_o=50$ and $N_i=25$, respectively.
    The final LSTM output is flattened and passed through two fully
    connected (FC) layers that parameterize a Gaussian latent distribution
    $\mathcal{N}(\mu, \sigma)$, where the mean $\mu$ and standard deviation $\sigma$
    are produced by separate FC layers.
    The dimensionality of the latent space, $N_{LS}$, is treated as a free
    hyperparameter.

    To enable gradient-based optimization, the reparameterization trick is applied to
    sample a latent embedding vector from the learned distribution:
    \begin{equation}
    z = \mu + \sigma \odot \epsilon, \qquad \epsilon \sim \mathcal{N}(0, I),
    \label{eq:reparam_trick}
    \end{equation}

    where $\epsilon$ is a noise variable and $\odot$ represents element-wise
    multiplication.

    The sampled latent vector $z$ is then provided as input to the decoder, which
    mirrors the encoder architecture.
    Specifically, $z$ is first projected through a fully connected layer, two
    concatenated LSTM layers, now with $N_i=25$ and $N_o=50$ respectively, and a
    final fully connected layer projecting from $\mathcal{R}^{N_o \times N_T}$
    space back to the $\mathcal{R}^{N_C \times N_T}$ EEG channel space.

    The LSTMVAE is trained with the ELBO loss $\mathcal{L}_\beta$ and we report the
    results obtained with $\beta=10^{-7}$.
    We applied dropout after each layer with a probability of $p=0.1$.

    \subsubsection{Convolutional Variational Autoencoder}

    We adapted the VAEEG architecture proposed by Zhao et al. \cite{Zhao2024},
    generalizing it to accommodate EEG segments of arbitrary number of input channels
    $\in \mathcal{R}^{N_C \times N_T}$.
    While the encoder structure remains consistent with the original
    implementation, modified only to accept multi-channel input, the decoder was
    adjusted to ensure signal reconstruction occurs directly in the channel space.

    The VAEEG utilizes a 1D Convolutional Neural Network (CNN) backbone specifically
    optimized for temporal feature extraction from EEG signals.
    The encoder consists of an initial series of CNN modules designed to capture
    features across varying observation scales, followed by four hierarchical blocks.
    Each hierarchical block integrates a single convolutional layer with two
    residual blocks, an arrangement designed to mitigate gradient degradation and
    enhance the generalization of learned representations.
    To ensure numerical stability and introduce non-linearity, every convolutional
    layer is followed by a normalization layer and a Rectified Linear Unit (ReLU)
    activation function.

    The encoding process terminates by mapping high-level temporal features to a latent
    distribution parameterized by mean ($\mu$) and standard deviation ($\sigma$)
    vectors.
    Again, the reparameterization trick from equation~\ref{eq:reparam_trick} is applied
    to enable backpropagation of gradients to the encoder.

    To facilitate signal recovery, the decoder is implemented as a symmetrical mirror of
    the encoder structure.
    Notably, we omit the task-specific \textit{tail} layer present in the original
    implementation; this allows the model to project the latent embeddings back into
    the original $\mathcal{R}^{N_C \times N_T}$ EEG channel space for full-signal
    reconstruction.

    As with the LSTMVAE, the VAEEG is trained with the ELBO loss $\mathcal{L}_\beta$,
    reporting here the results obtained with $\beta=10^{-7}$.

    \subsubsection{Transformer Autoencoder}

    We further evaluated a hybrid CNN-Attention architecture, also based on the
    framework defined by Zhao et al. \cite{Zhao2024} and
    generalized here for multi-channel EEG processing.

    As with the previous architectures, the encoder was adapted to handle an
    arbitrary number of input channels, while the decoder was modified by omitting
    the task-specific \textit{tail} layer to facilitate direct reconstruction in the
    channel space.

    The initial encoding stage consists of a multi-channel 1D CNN with varying
    kernel sizes, acting as a tokenization layer.
    This stage expands the feature (channel) dimension and captures multi-scale local
    temporal filters.
    Following a transposition of the temporal and channel dimensions, the sequence
    is processed by two Transformer encoder blocks.
    Each block is composed of a multi-head self-attention mechanism with three heads
    and a 64-node position-wise feed-forward propagation layer.

    The resulting feature map is flattened and projected through a Multi-Layer
    Perceptron (MLP) into a $N_{LS}$-dimensional latent space.
    For the reconstruction phase, this model adopts the same mirrored convolutional
    decoder architecture used in the VAEEG.
    By utilizing a common decoder across both convolutional and attentional models,
    we ensure that any observed performance variations are strictly attributable to the
    encoding mechanisms rather than differences in the generative process.

    Following the original study's implementation, the Transformer architecture does
    not follow the typical VAE structure, but instead a classical autoencoder
    architecture and is therefore trained with the reconstruction MSE loss
    $\mathcal{L} = MSE(x, \hat{x})$~\cite{Zhao2024}.

\subsection{Network Training}
    All EEG segments (synthetic and empirical) utilized during the training and
    evaluation of the models consisted of two-second long segments ($f_s=250$ Hz)
    in a 19-channel configuration, except for the Spectral Learning Order study described below,
    where single-channel segments were used, mimicking the approach by Rahaman et al.
    \cite{Rahaman2018}.

    Implemented networks followed the same training procedure and hyperparameters.
    Network convergence has been ensured by visual inspection of the learning curves,
    visualizing the evolution of both the training and validation losses.

    All networks are trained with default weight initialization, for
    $N=500$ epochs.
    Backpropagation of the training loss was used for weight updating via the Adam
    optimizer~\cite{Kingma_adam} with an initial learning rate of
    $\eta = 5\times 10^{-4}$ and a Cosine Annealing learning rate scheduler with
    $T_{\max}=10$ training epochs.
    A batch size of 128 has been used for all studies.
    Early stopping based on the validation loss was used, saving the model weights
    with the lowest validation loss. All reported results belong to the  held-out test partition.

    To ensure the statistical robustness of our findings and mitigate the influence of stochastic initialization, each model was trained on the synthetic data across 10 independent iterations using unique random seeds.
    The results reported herein represent the mean performance metrics across these runs, effectively accounting for variability inherent in the training process.
    When training on empirical data, only a single seed was used, due to the compute constraints arising from the dataset's size.

\subsection{REVE Foundation Model}

To complement our investigations on application-specific autoencoders, we also considered REVE (Representation for EEG with Versatile Embeddings)~\cite{elouahidi2025reve}, a large-scale self-supervised foundation model for EEG. REVE is a Transformer
encoder pretrained with a masked-autoencoding objective on over $60{,}000$ hours
of recordings drawn from $92$ datasets and $25{,}000$ subjects, and is explicitly
designed to generalize across heterogeneous acquisition montages. Each electrode
time series is partitioned into temporal patches that are linearly projected into
tokens; a four-dimensional positional encoding jointly represents the
three-dimensional spatial coordinates of every electrode and the temporal index
of each patch, allowing the model to ingest recordings with an arbitrary number
of channels $N_C$ and arbitrary duration $N_T$.
During pretraining, a primary objective reconstructs the raw masked signal
directly, while a secondary objective reconstructs the input from a single global
token obtained via attention pooling over all channel--patch tokens; both terms
minimize an $L_1$ reconstruction loss. We employ the publicly released
\texttt{reve-base} and \texttt{reve-large} checkpoints.

Unlike the VAEEG and LSTMVAE, REVE is not trained on our synthetic or empirical data: we use it strictly
as a frozen feature extractor, keeping all weights fixed and performing no
fine-tuning, so that our analysis probes the quality of the pretrained
representation itself rather than a task-specific adaptation. Each
$N_C = 19$-channel, $2$\,s epoch is resampled to REVE's native $200$\,Hz sampling
rate and passed through the encoder together with the corresponding electrode
coordinates. We summarize the resulting channel-patch token grid into a single
$d = N_{LS}$-dimensional embedding ($N_{LS}=512$ for \texttt{reve-base} and $N_{LS}=1216$ for \texttt{reve-large}) using the model's native attention-pooling (global) token, and treat this vector as the latent representation, which is subsequently analyzed with the same downstream pipeline applied to the VAE latent spaces.

\subsection{Latent Space Exploration}\label{sec:LSexploration}
    \subsubsection{UMAP-based visualization}
    We used dimensionality reduction techniques to visualize the structure of the
    learned latent space.
    Specifically, we employed Uniform Manifold Approximation and Projection (UMAP),
    a nonlinear manifold learning algorithm that constructs a low-dimensional representation
    by preserving both local neighborhood relationships and, to a lesser extent,
    global data structure~\cite{McInnes2018}.

    UMAP was applied to the latent embeddings extracted from any of the trained
    networks, with each point in the two-dimensional space corresponding to a single
    multi-channel EEG segment.
    Points are colored according to features of interest, such as diagnostic label (HC,
    SCD, MCI, AD) or relative band power (theta, alpha, etc.).

    This projection enabled qualitative inspection of the spatial organization, overlap,
    and separation of embeddings, thus providing a qualitative and intuitive method of
    network training evaluation.

    All UMAP projections were computed using the same parameter settings to ensure
    consistency across visualizations.
    We made use of the \textit{umap-learn} Python package~\cite{McInnes2018_method} with
    all the default parameters except we fixed \textit{n\_neigbors=100}.

    \subsubsection{Offline Non-Linear Classifier Accuracy}
    To evaluate the quality of the learned representations and ensure the latent space
    effectively encodes diagnostic features, we implemented an offline non-linear
    classification task.
    This serves as a downstream validation of the unsupervised feature distillation: if the
    autoencoder has successfully organized the latent space, a simple classifier should be
    able to predict subject diagnosis using only the learned embeddings.

    After the training of each autoencoder architecture, we extracted the latent
    embeddings $\mathbf{z}$ for all EEG segments across the training, validation, and
    testing partitions.
    These partitions remained identical to those used during the autoencoder training phase
    to prevent data leakage.

    The training set embeddings were used to optimize the classifier, the validation set for
    early stopping, and the held-out test set for reporting final accuracy metrics.

    The non-linear classifier is a fully connected Multi-Layer Perceptron (MLP) that accepts
    latent vectors of dimension $N_{LS}$ and forwards it through two 64-unit layers
    with ReLU activations.

    A final linear layer generates class probabilities; the network's prediction is then
    determined by the index of the maximum probability value.

    The classifier is trained with a CrossEntropy loss and stochastic gradient
    descent for 500 training epochs and with a learning rate of 0.01.

    % \subsubsection{Cluster Validity Measures}
    %
    % Cluster validity measures are statistical techniques designed to assess the quality of
    % clustering solutions by quantifying how well samples assigned to the same cluster are
    % separated from samples belonging to other clusters.
    % Typically, these metrics evaluate whether within-cluster similarity is high and
    % between-cluster similarity is low.
    %
    % In the present work, such measures are repurposed to assess the structure of the
    % autoencoder latent space.
    % Rather than evaluating the output of an explicit clustering algorithm, we use the
    % Silhouette score to quantify how well latent embeddings corresponding to the same
    % clinical label (HC, SCD, MCI, AD) group together and how distinctly they are separated
    % from embeddings associated with other diagnostic categories.
    % In this sense, the Silhouette score provides a measure of class-wise compactness and
    % separability in the learned latent space.
    %
    % In particular, the Silhouette Score~\cite{Rousseeuw1987} is calculated using the
    % mean intra-cluster distance $a_i$ and the mean nearest-cluster distance $a_i$ for each
    % sample $i$, such that:
    % \begin{equation}\label{eq:silhouette}
    %     s(i) = \frac{b_i - a_i}{max(a_i, b_i)}, S = \sum_i s(i)
    % \end{equation}
    %
    % Following the implementation in \textit{scikit.learn}, we report the mean Silhouette
    % score over all samples $S$.
    % The Silhouette score's range is between -1 and 1, with 1 representing the best
    % possible
    % clustering and 0 representing overlapping clusters.

\subsection{Spectral Learning Order}\label{sec:methods_spectral_bias}

    In their seminal work on spectral bias, Rahaman and colleagues present a synthetic
    regression problem to demonstrate that MLPs prioritize learning lower
    frequency mappings, regardless of their relative~\cite{Rahaman2018}.
    Here, we adapt the methodology of their ``Synthetic Experiment 1`` to inspect the learning dynamics of the autoencoders investigated in this study.

    We generated a synthetic dataset of 500 samples (split into 200/200/100 for
    training, validation, and testing, respectively) consisting of two-second-long
    single-channel signals.
    Each signal $x_n(t)$ is generated according to the following equation:

    \begin{equation}\label{eq:spectral_bias_data}
    x_n(t) = \sum_{i=1}^{10} A_i \sin(2\pi f_i t + \phi_i + \phi_n),
    \end{equation}

    where the frequency set is defined as $f \in \{5, 10, 15, \dots, 50\}$~Hz. For
    each frequency $f_i$, a fixed phase $\phi_i$ is randomly sampled from
    $\mathcal{U}(0, 2\pi)$.
    To generate the $N=500$ unique samples, an additional sample-specific phase
    $\phi_n$ is obtained from $\mathcal{U}(0, 2\pi)$.

    We evaluate two distinct amplitude configurations.
    In the first case, all amplitudes are equal ($A_i=1$) to isolate the effect of
    frequency on learning speed.
    In the second case, $A_i$ follows a pink-noise-like decay, defined as
    $A_i = 1 / f_i^\chi$ with $\chi=0.75$.
    This latter configuration provides a more ecologically valid representation of
    the power spectral density (PSD) typically observed in EEG signals.

    During autoencoder training, we monitored the spectral evolution of the
    reconstructions.
    We computed the Fourier Transform of the network output, $FT_{out}(f)$, and
    compared it to the Fourier Transform (FT) of the target sample, $FT_{target}(f)$, via
    their ratio.
    Under this metric, a frequency $f_j$ is considered successfully represented when
    the ratio $\frac{FT_{out}(f_j)}{FT_{target}(f_j)} \approx 1$.
    Conversely, a ratio near 0 indicates that the frequency has not yet been
    captured by the model. 

While we are heavily inspired by Rahaman et al.'s work on spectral bias, it is worth clarifying a conceptual difference in what our metric captures. Rahaman et al. characterise the frequency content of the function the network computes over its input domain, showing that low-frequency components of the target mapping are fit before high-frequency ones. In their scalar regression setting, this function-space frequency coincides with the spectral content of the target signal. In our autoencoding setting, the two are not identical: the network maps an entire signal window to its reconstruction, and the frequency axis of 
$FT_{out}(f)$ refers to the spectral content of the reconstructed signal rather than to the frequency composition of the encoder or decoder as functions over their respective input domains. Because our metric tracks the order in which frequencies are recovered during training, we name this observation \textit{Spectral Learning Order}, which we consider analogous but not equivalent to the spectral bias described in \cite{Rahaman2018}.

\section{Results}\label{sec:results}

\subsection{Discriminating higher frequency alterations requires larger latent spaces}

    \begin{figure}[h!]
        \centering
        \includegraphics[width=\textwidth]{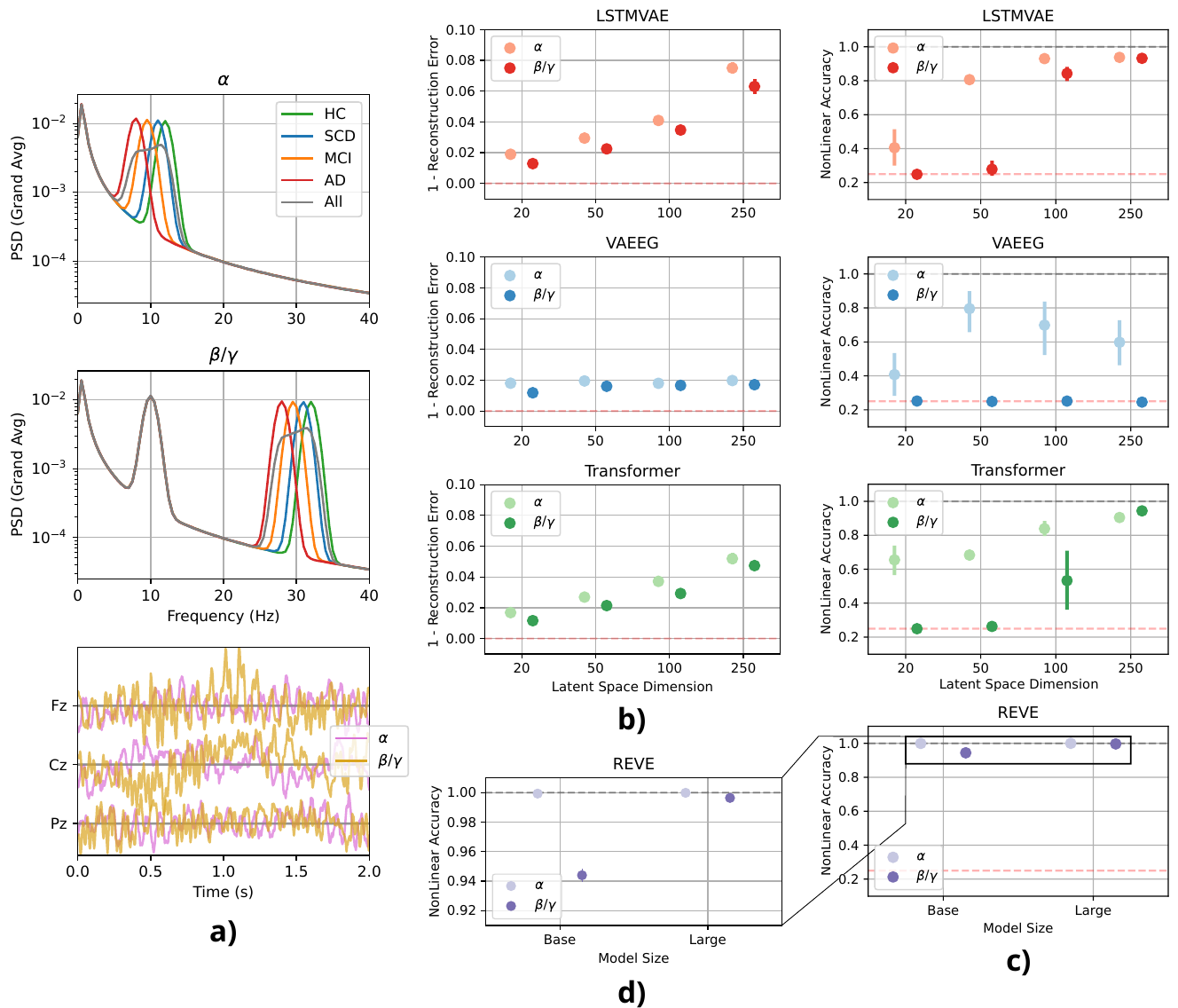}
        \caption{
        \textbf{Performance Evaluation on Synthetic EEG Datasets.}
        \textbf{\textbf{(a) Characterization of Synthetic Datasets}:} The upper (middle) panels display the group-specific Power Spectral Density (PSD), averaged across channels and subjects within the validation partition for the $\alpha$ ($\beta / \gamma$) datasets. The lower panels show representative 2-second EEG segments for the three simulated midline channels across both datasets.
        \textbf{\textbf{(b) Reconstruction Fidelity across Latent Dimensions}:} Model performance is reported as $1 - \text{MSE}$ (where 1 indicates perfect reconstruction) on the held-out test partition. Results are shown for the LSTMVAE (upper), VAEEG (middle), and Transformer (lower) architectures as a function of the latent space dimensionality ($N_{LS}$) and dataset type.
        \textbf{\textbf{(c) Latent Space Discriminative Power}:} Classification accuracy on the held-out test partition of a non-linear MLP trained to predict diagnostic group (HC, SCD, MCI, AD) from the latent embeddings. Accuracy is reported for the LSTMVAE, VAEEG, Transformer, and REVE (top to bottom), across varying latent dimensions. \textbf{(d)} provides a detail-view of REVE's high accuracy. 
        Theoretical chance accuracy (25\%) is denoted by the horizontal red dashed line.
        Error bars indicate 95\% confidence interval. 
        }\label{fig:synth_data_perf}
    \end{figure}

    We trained three autoencoder networks of different characteristics (LSTMVAE,
    VAEEG, Transformer) on two different synthetic datasets simulating a cognitive
    decline spectrum EEG dataset.
    Figure~\ref{fig:synth_data_perf} displays the reconstruction error ($1 - MSE$)
    over the test partition produced by each network when increasing the latent space
    dimension ($N_{LS}$).
    As has been previously reported, increasing $N_{LS}$ generally enhanced
    reconstruction, except for the VAEEG architecture, where the improvement was
    almost non-existent, which may be a result of our very limited hyperparameter exploration, which was outside of the scope of this paper.
    While reconstruction error remained similar for a given network and latent space
    size, the $\alpha$ dataset typically achieved better reconstructions.

    Latent-space embeddings were used to classify each sample into one of four cognitive decline categories via an MLP, thus evaluating encoding quality. For this analysis, we additionally included embeddings generated by REVE with frozen weights.
    We report the MLP's accuracy over the testing partition in
    Figure~\ref{fig:synth_data_perf}.
    For the $\alpha$ dataset, alterations in the $\alpha$ band were consistently
    captured by the autoencoders with latent space dimension larger than $N_{LS}=50$.
    However, alterations in $\beta / \gamma$ required at latent space dimensions of
    at least $N_{LS}=100$ for the LSTMVAE and Transformer architectures to
    discriminate between-group differences, while the VAEEG network could not
    meaningfully distill the dataset's group differences independently of the latent
    space dimensions.
    Additionally, the VAEEG architecture suffered from a counter-intuitive
    degradation in the encoding of the $\alpha$ dataset's alterations with increasing
    $N_{LS}$, questioning whether some autoencoders with large latent
    spaces may be suitable for disentangling simple data generative factors.
    REVE generally displayed near perfect accuracy. However, the Base ($N_{LS}=512$) model displayed an average accuracy of 94.4\%, demonstrating the increased difficulty in accurately disentangling higher-frequency features with a limited latent size. Both LSTMVAE and Transformer architectures displayed similar MLP accuracies to REVE in their largest configurations.

    \begin{figure}[h!]
        \centering
        \includegraphics[width=\textwidth]{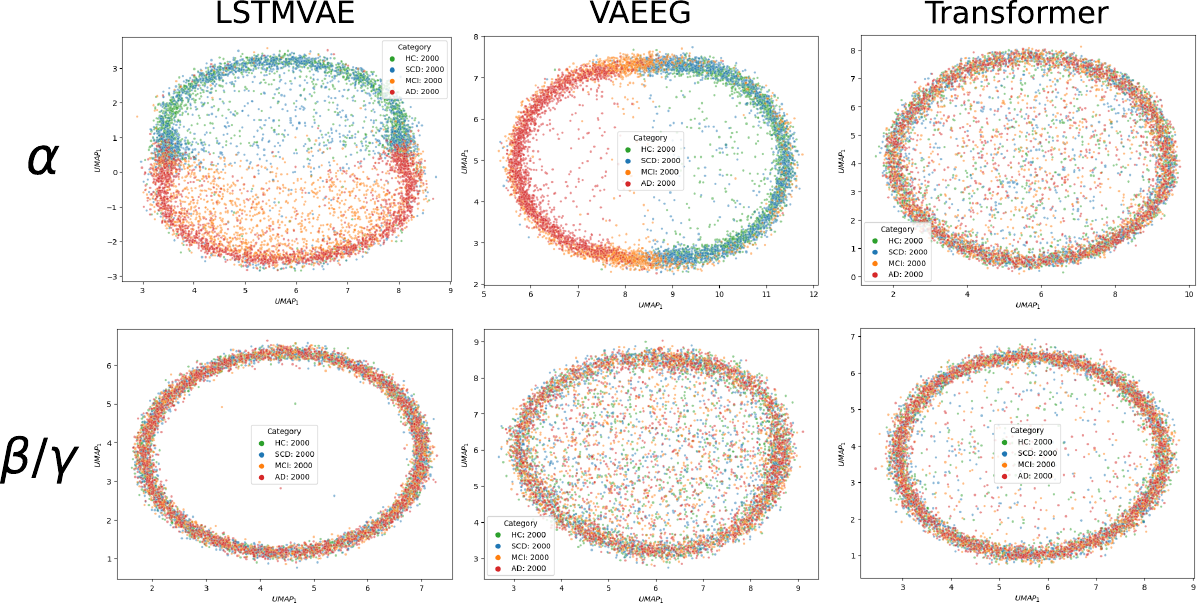}
        \caption{
            \textbf{UMAP Projections of Latent Space Embeddings ($N_{LS}=50$).}
            Two-dimensional UMAP visualizations of the latent manifolds learned by the
            LSTMVAE (left), VAEEG (middle), and Transformer (right) architectures. The
            upper row displays projections for the $\alpha$ slowing dataset, while the
            lower row represents the $\beta / \gamma$ slowing dataset. Each point
            corresponds to a multi-channel EEG segment from the held-out test partition,
            color-coded by the synthetic diagnostic group (HC, SCD, MCI, AD) to
            illustrate the degree of class-wise clustering and separability within
            the learned latent space.
        }
        \label{fig:synth_data_umaps_50}
    \end{figure}

    We additionally projected the latent space embeddings of the samples
    in the testing partition via UMAP for the three architectures with $N_{LS}=50$,
    reported in Figure~\ref{fig:synth_data_umaps_50}.
    The projections were able to visually capture the latent structure for both the
    LSTMVAE and VAEEG in the synthetic $\alpha$ dataset.
    Clear visual overlaps between the HC/SCD and MCI/AD classes can be observed, which
    is in accordance with the closer frequency peaks between these classes as has
    been described in the methods, and can explain the misclassifications of the
    non-linear classifiers.
    % Additionally, same pattern is observed when LS is over-regularized by ELBO and
    % we have only gaussian distribution, something of note.

\begin{figure}[h!]
        \centering
        \includegraphics[width=\textwidth]{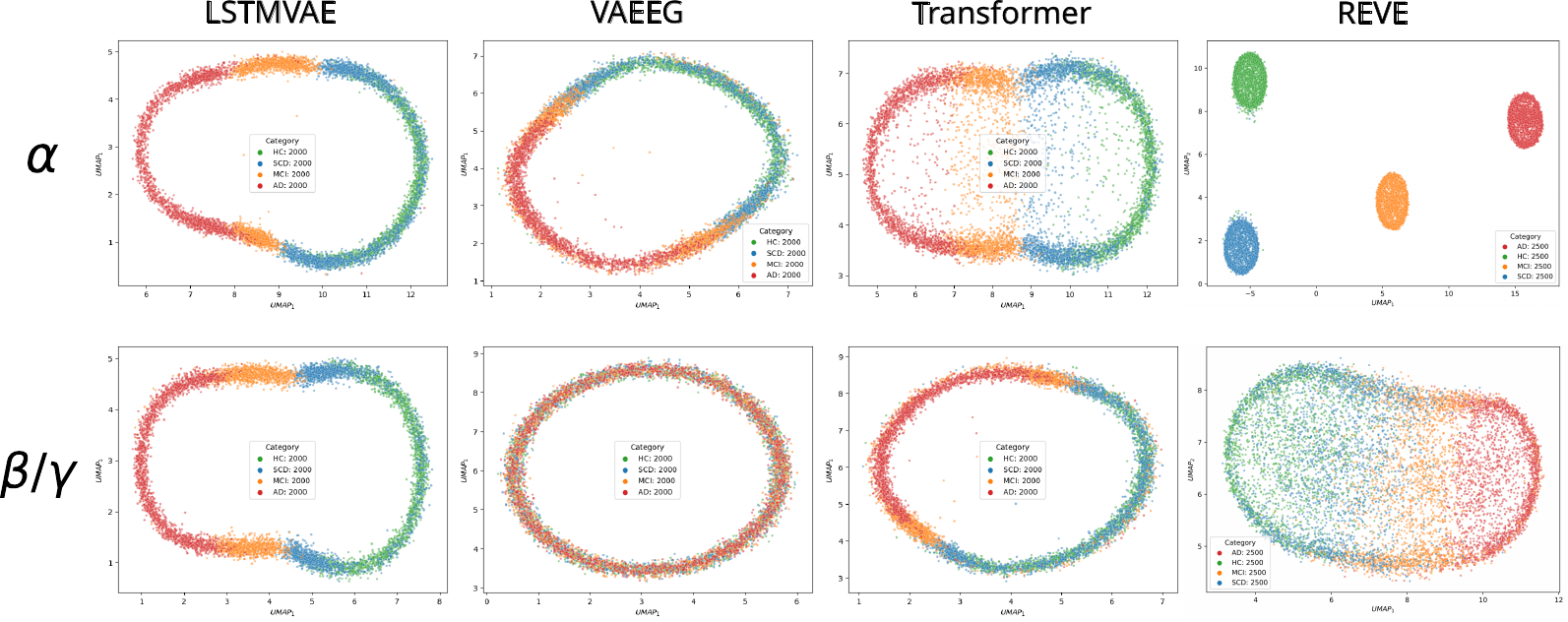}
        \caption{
        \textbf{UMAP Projections of Latent Space Embeddings ($N_{LS}=250$).}
        Two-dimensional UMAP visualizations of the latent manifolds learned by the (left to right)
        LSTMVAE, VAEEG, Transformer, and REVE architectures. The
        upper row displays projections for the $\alpha$ slowing dataset, while the
        lower row represents the $\beta / \gamma$ slowing dataset. Each point
        corresponds to a multi-channel EEG segment from the held-out test partition,
        color-coded by the synthetic diagnostic group (HC, SCD, MCI, AD) to
        illustrate the degree of class-wise clustering and separability within
        the learned latent space.
        }
        \label{fig:synth_data_umaps_250}
    \end{figure}

    In the $\beta / \gamma $ case, UMAP projections did not elucidate any
    meaningful structure in the LS, which further corroborate the lack of feature
    distillation for this dataset at $N_{LS}=50$ previously accounted by the low
    accuracy of the non-linear classifier.
    Figure~\ref{fig:synth_data_umaps_250} shows how structure could be recovered for
    the higher-frequency dataset in the $N_{LS} = 250$ configuration of the LSTMVAE, Transformer, and REVE models, again following the non-linear classifier accuracy trend. We choose to display REVE's projections in the supplementary Figure as the latent size of the Base model better matches $N_{LS}=250$.
    Finally, despite displaying high non-linear classifier accuracy, the Transformer
    latent space projections lacked structure.
    Although this may be a result of the UMAP algorithm not being able of disentangling
    the latent structure, one should note that Transformer was trained with an MSE
    loss instead of an ELBO loss, thus skipping latent space regularization,
    resulting in a less structured, albeit still informative, embedding function.

\subsection{Spectral Content Predominates Over Spectral Learning Order in EEG Reconstruction}

    \begin{figure}[h!]
        \centering
        \begin{subfigure}{\textwidth}
            \centering
            \includegraphics[width=\textwidth]{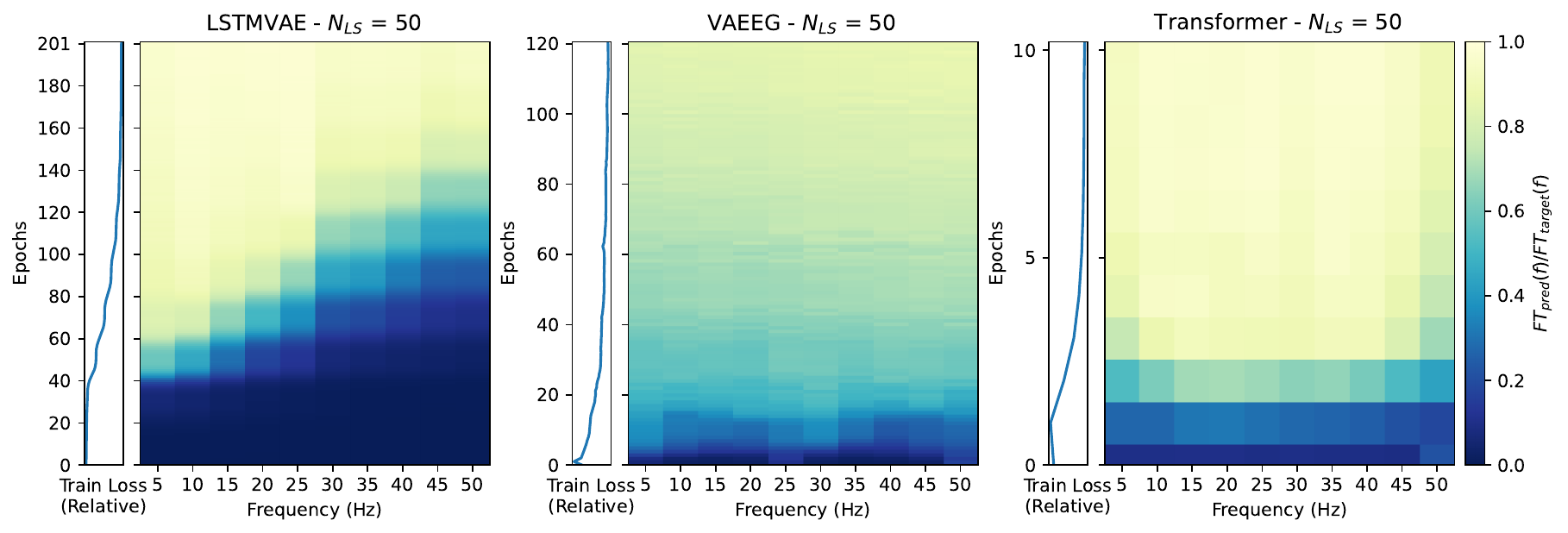}
        \end{subfigure}

        \begin{subfigure}{\textwidth}
            \centering
            \includegraphics[width=\textwidth]{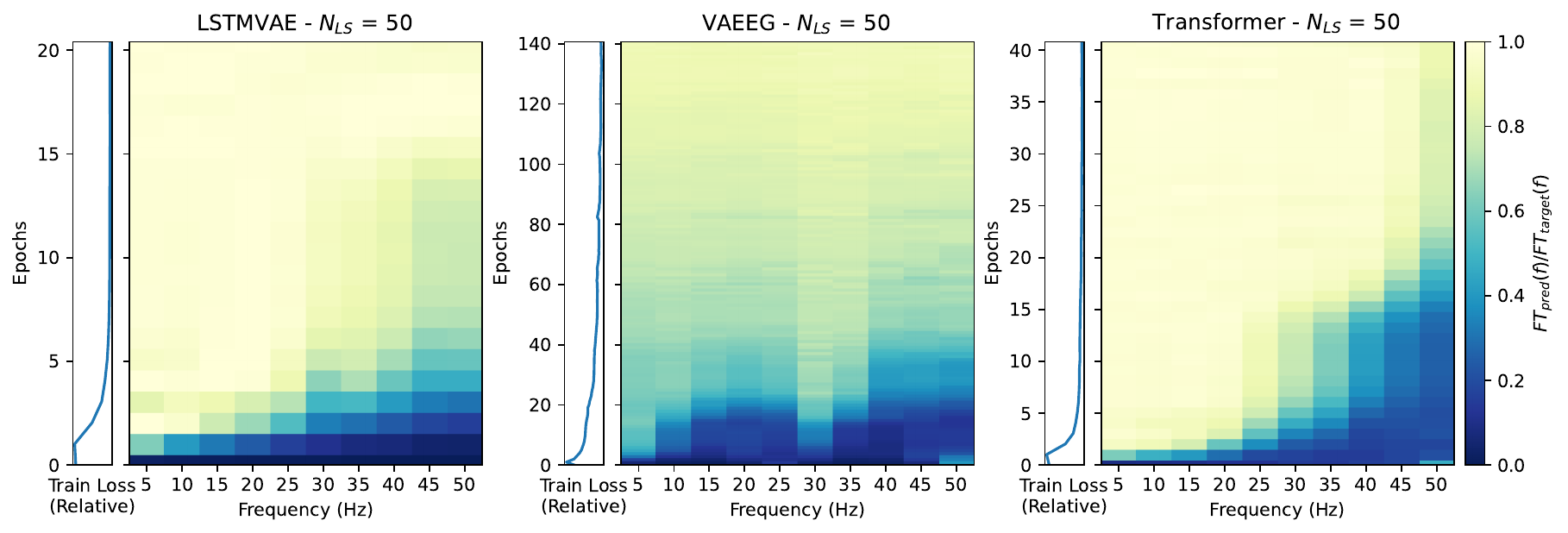}
        \end{subfigure}

        \caption{
            \textbf{Spectral Learning Order} Synthetic experiment adapted
            from Rahaman et al. \cite{Rahaman2018} illustrating the evolution of
            frequency learning as a function of training progress. The heatmaps
            represent spectral fidelity, defined as the ratio between the
            reconstructed and target amplitudes for each frequency component
            ($f \in \{5, 10, \dots, 50\}$ Hz). A ratio of 1 (yellow) indicates
            successful reconstruction (hence encoding), while a ratio of 0 (blue) indicates the component
            has not yet been captured.
            Two scenarios are compared across the three architectures (LSTMVAE, VAEEG,
            and Transformer) with $N_{LS} = 50$: the equal-amplitude condition
            ($A_i=1$, upper panels), isolating architectural bias; and the
            decaying-amplitude condition ($A_i \propto 1/f_i^{0.75}$, lower panels),
            simulating the power distribution of empirical EEG to observe
            magnitude-driven masking effects.
        }
        \label{fig:spec_bias_50}
    \end{figure}

    We investigated the impact of spectral learning order to determine whether it
    may behave as a bottleneck for feature discovery in EEG autoencoders.
    By monitoring the epoch-wise reconstruction fidelity of individual frequency
    components ($f_i$), we evaluated how both architectural design and signal power
    distributions influence the learning trajectory.
    These dynamics were captured under two distinct scenarios: a controlled
    equal-amplitude condition ($A_i=1$) and a decaying-amplitude condition
    ($A_i \propto 1/f_i^{0.75}$) mimicking the power spectral density of human EEG.

    Under the equal-amplitude condition (Figure~\ref{fig:spec_bias_50}, upper), our
    results revealed a stark contrast between recurrent and non-recurrent architectures.
    The LSTMVAE exhibited a significant spectral learning order, consistently reconstructing
    low-frequency oscillations in the early stages of training while only capturing
    higher frequencies as the model converged.
    Conversely, both the VAEEG (CNN-based) and the Transformer-based network showed no
    such ordering, capturing high-frequency components near-simultaneously with lower ones.

    However, when the networks were trained on signals featuring a $1/f^{0.75}$
    amplitude decay (Figure~\ref{fig:spec_bias_50}, lower), all three architectures exhibited a spectral learning order, prioritizing the
    reconstruction of low-frequency components first.

    These findings indicate that, while certain autoencoder architectures are inherently more capable of capturing high-frequency information from the onset of training, the power distribution of EEG signals imposes a ``pseudo-spectral bias`` across all models.

We extended our spectral learning order study to REVE's specific architecture and training objective, detailed in Appendix \ref{sec:app_reve_spec_bias}. Unlike the generic autoencoders discussed above, REVE shows no evidence of spectral ordering regardless of spectral content (Figure \ref{fig:spec_bias_REVE}). Our extended analyses identify the linear patch tokenization as the most likely factor underlying this robustness. Nonetheless, REVE's latent space shows worse discriminative power on the $\beta/\gamma$ than on the $\alpha$ dataset(Figure \ref{fig:synth_data_perf}). This apparent discrepancy reflects a difference in what each analysis captures: the spectral learning order experiment measures learning dynamics (when each frequency is acquired by the reconstruction pathway) whereas the classification gap highlights the frequency-resolution of the converged model's embeddings when pretrained on real EEG. Because empirical EEG exhibits a 1/f power spectrum, REVE's amplitude-weighted L1 objective seems to allocate representational capacity in proportion to signal power, yielding coarser resolution at higher frequencies than at lower ones.
    
    For the application of autoencoders and foundation models to raw EEG, this implies that the spectral content (the relative power of different frequencies) is a more critical determinant of learning dynamics and representational resolution than the architectural spectral biases of networks themselves.

\subsection{Autoencoders capture meaningful EEG features but suffer from spectral attenuation}

    \begin{figure}[h!]
        \centering
        \includegraphics[width=\textwidth]{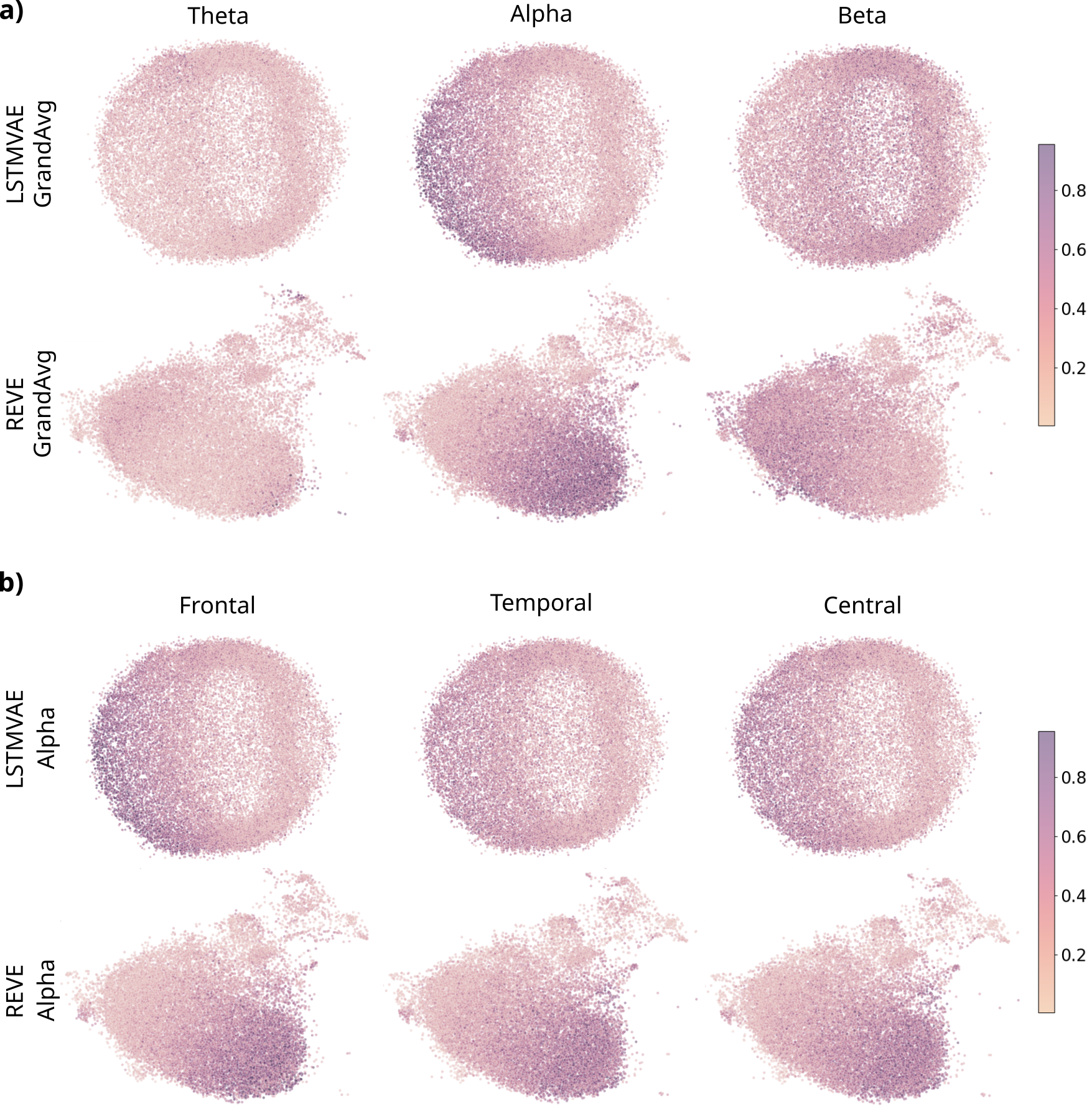}
        \caption{
            \textbf{Spectral Organization of the LSTMVAE Latent Space for the held-out test partition of the BBHI Dataset.}
            \textbf{(a)} UMAP projections of the latent embeddings color-coded
            by the grand-average (average over all channels) relative power within standard EEG frequency bands:
            theta ($4-8$ Hz), alpha ($8-13$ Hz), and beta
            ($13-30$ Hz).
            \textbf{(b)} Topographically resolved UMAP projections of alpha relative band power for the LSTMVAE (upper) and REVE (lower) models. Color-coding alpha band power is the result of averaging alpha band power over different sets of channels: Frontal (F7, F3, Fz, F4, F8), Temporal (T7, T8), and Central
            (C3, Cz, C4).
            These projections aim to illustrate the emergence of spectral and spatial
            representational structure within the unsupervised latent manifold.
        }
        \label{fig:bbhi_umaps}
    \end{figure}

    To corroborate our synthetic findings with empirical data, we trained the reconstruction autoencoders on a large-scale EEG dataset comprising $N=943$ subjects. In parallel, we used REVE to generate embeddings for the same dataset. Figure \ref{fig:bbhi_umaps} shows UMAP projections of the latent embeddings, where each point is the embedding for each 2-second-long multi-channel EEG window, coloured by the window's relative band power. 
    
    Both LSTMVAE and REVE demonstrated a clear capacity to capture fundamental EEG features and
    recover established domain knowledge, specifically regarding the canonical alpha ($8-13$ Hz) rhythm. However, this organized representational structure was less evident for the theta (4–8 Hz) and beta (13–30 Hz) bands. In particular, LSTMVAE exhibited only marginal structure in the beta band, while REVE appeared to display a structure complementary to that observed in the alpha band.
    
    Given that the alpha band typically exhibits the highest power in the human EEG
    spectrum, its successful reconstruction confirms our earlier observations:
    autoencoders trained to minimize raw signal reconstruction error are inherently
    biased toward capturing frequency components with the largest contribution to
    the total signal variance.
    This behavior reinforces the ``pseudo-spectral bias`` identified in our synthetic
    experiments.

    Thus, latent space dynamics appear to prioritize the dominant rhythms in the
    signals, resulting in meaningful encodings for low-frequency features.
    In contrast, frequency components with smaller power contributions, such as
    higher beta and gamma bands, do not show distinct representations in
    the latent space.
    
    Despite the successful recovery of global temporal rhythms, we found no qualitative evidence of spatially-specific representations: the latent representations did not significantly reflect the topological or sensor-specific nuances of the 19-channel input. This absence may simply be an artifact of the qualitative nature of the UMAP embeddings, and spatial information may still be encoded in the latent space. Nonetheless, it is clear that the dominant axis of representational structure is frequency/temporal, rather than spatial.

\begin{figure}[h!]
        \centering
        \begin{subfigure}{0.3\textwidth}
            \centering
            \includegraphics[width=\textwidth]{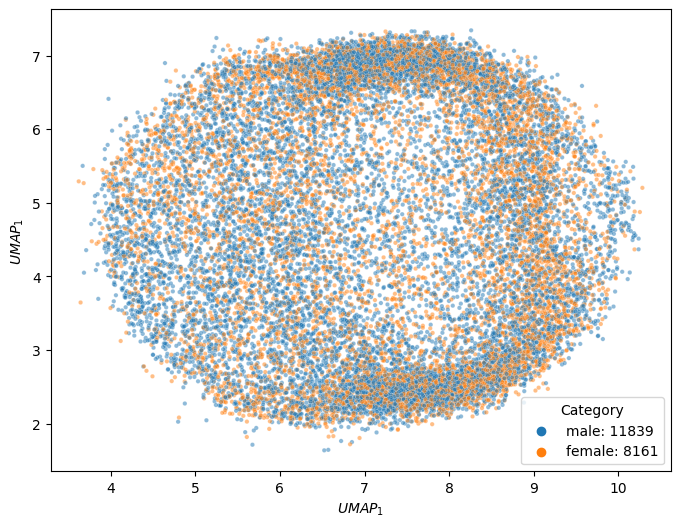}
            \caption{Sex}
        \end{subfigure}
        \begin{subfigure}{0.3\textwidth}
            \centering
            \includegraphics[width=\textwidth]{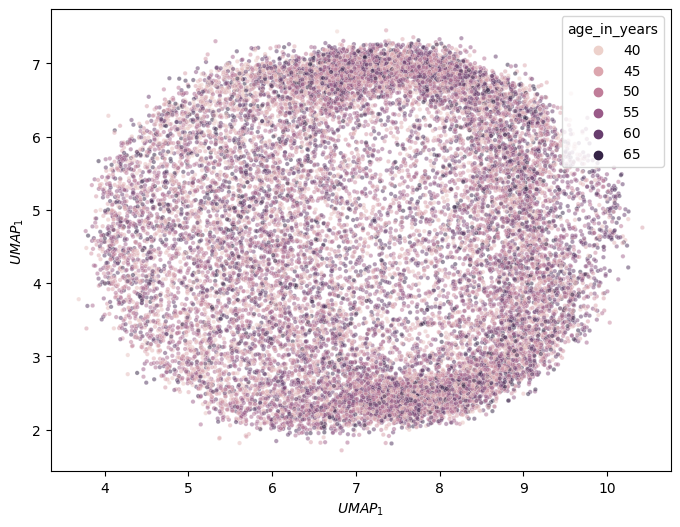}
            \caption{Age}
        \end{subfigure}
        \begin{subfigure}{0.3\textwidth}
            \centering
            \includegraphics[width=\textwidth]{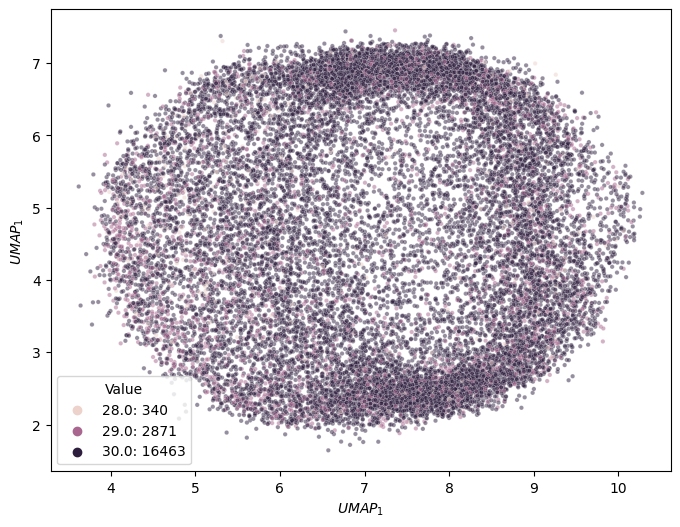}
            \caption{MMSE}
        \end{subfigure}
        \caption{UMAP projection of LS embeddings for LSTMVAE $N_{LS}=50$, coloured by
        demographic information (sex, age, and cognitive performance measured with MMSE)}
        \label{fig:umap-demographics-lstmvae}
    \end{figure}

\begin{figure}[h!]
        \centering
        \begin{subfigure}{0.3\textwidth}
            \centering
            \includegraphics[width=\textwidth]{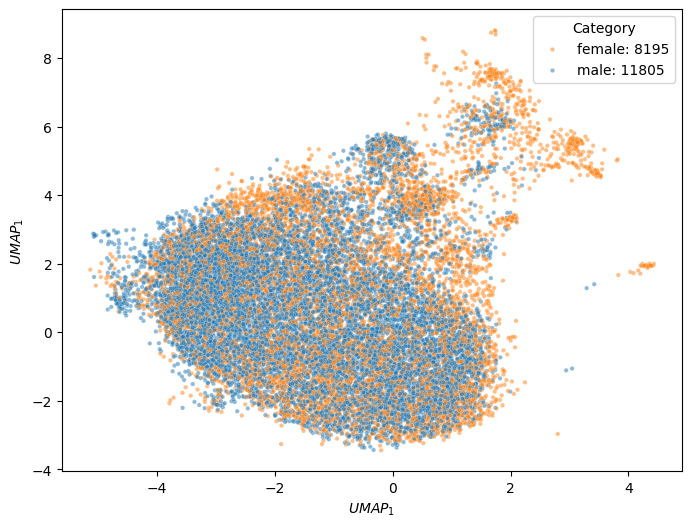}
            \caption{Sex}
        \end{subfigure}
        \begin{subfigure}{0.3\textwidth}
            \centering
            \includegraphics[width=\textwidth]{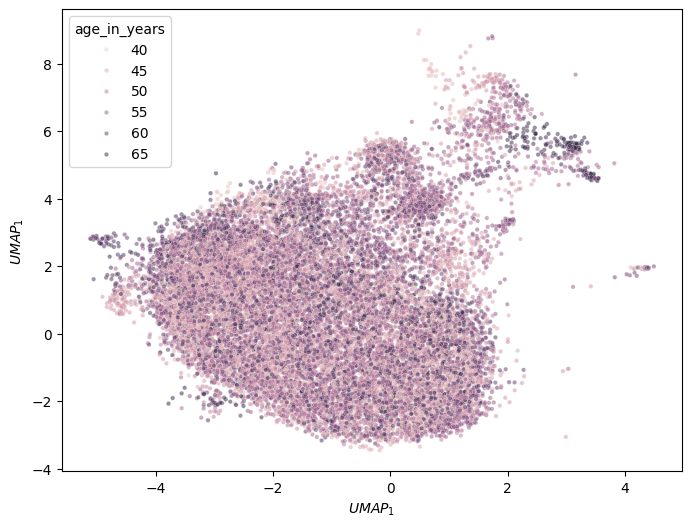}
            \caption{Age}
        \end{subfigure}
        \begin{subfigure}{0.3\textwidth}
            \centering
            \includegraphics[width=\textwidth]{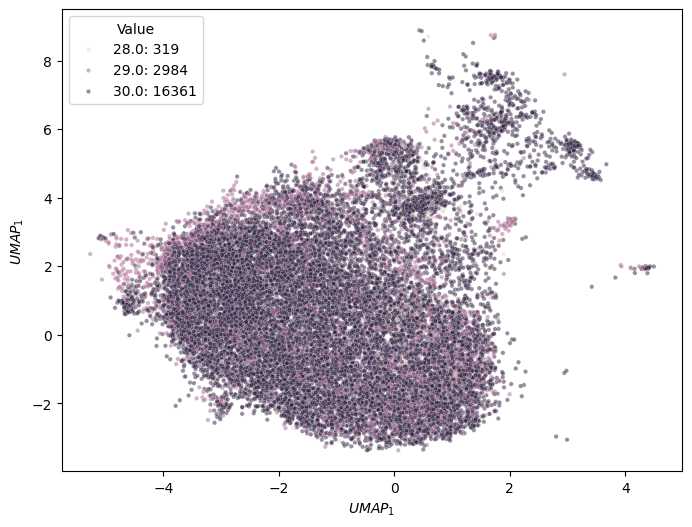}
            \caption{MMSE}
        \end{subfigure}
        \caption{UMAP projection of LS embeddings for REVE Base $N_{LS}=512$, coloured by
        demographic information (sex, age, and cognitive performance measured with MMSE)}
        \label{fig:umap-demographics-reve}
    \end{figure}

    While Figure~\ref{fig:bbhi_umaps} illustrates that the models capture
    discernible frequency patterns within the EEG data, these representations do not appear to align with clinical or cognitive metrics.
    As shown in Figure~\ref{fig:umap-demographics-lstmvae} and Figure\ref{fig:umap-demographics-reve}, we observed no significant
    correlation between the latent space distributions and sex, age, or
    cognitive performance (Mini-Mental State Examination, MMSE, score), for either the LSTMVAE or the REVE model. It is worth noting, however, that clinically relevant EEG features are typically averaged over all windows in a recording, so expecting a clear correlation between demographics and single epochs is unwarranted. Merging a subject's per-epoch embeddings into a single feature vector for later use in demographic prediction would be the natural next step.\\

    % This disconnect suggests that while the autoencoder is an effective tool for
    % signal compression and the recovery of dominant spectral features, the features
    % it prioritizes for reconstruction do not necessarily overlap with the biomarkers
    % required for clinical differentiation or the assessment of cognitive decline.

    \begin{figure}[h!]
        \centering
        \includegraphics[width=\textwidth]{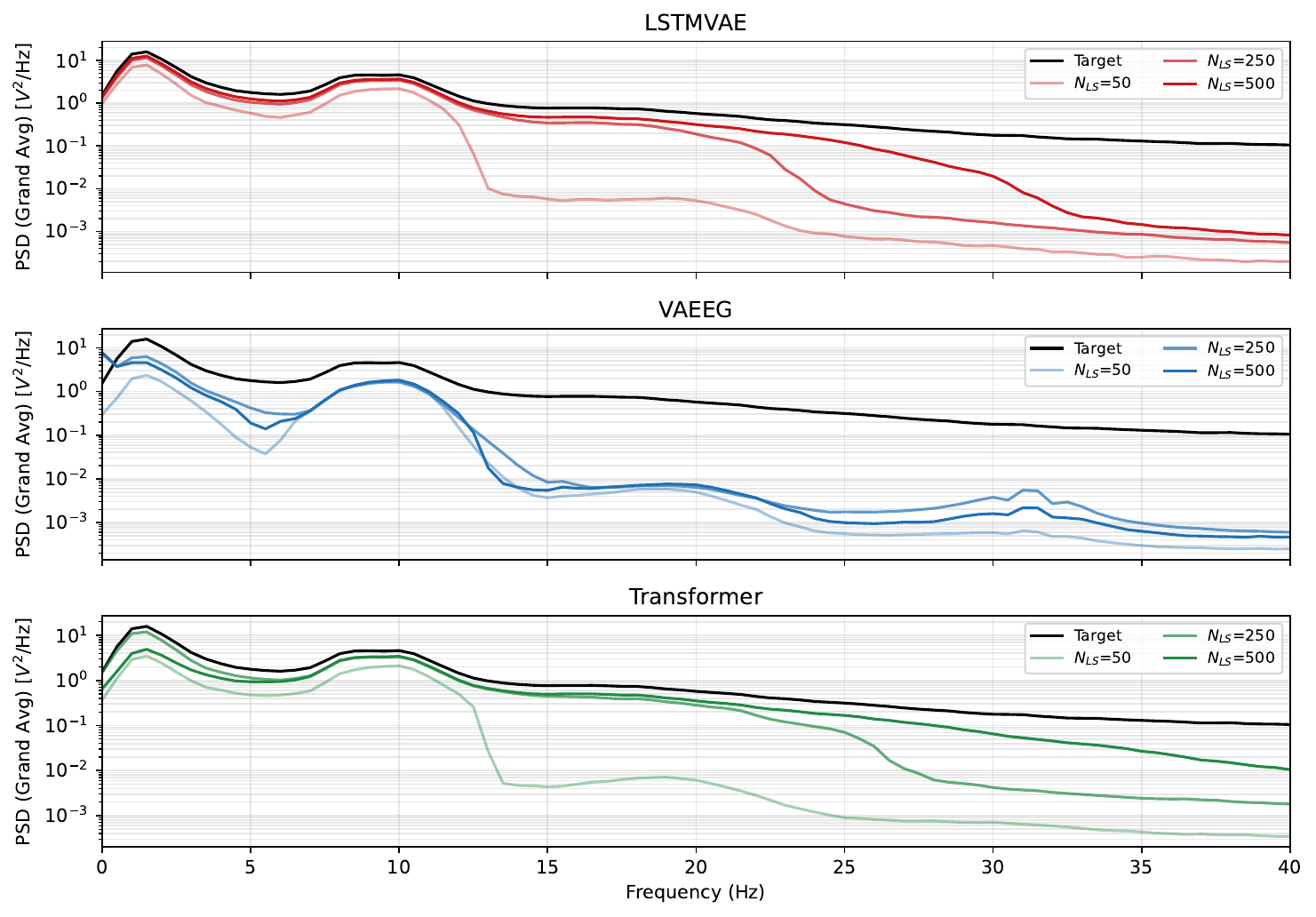}
        \caption{
            \textbf{Frequency-resolved spectral decay across reconstruction autoencoders as a function of latent space dimension ($N_{LS}$), evaluated on the held-out test partition of the BBHI dataset.} Power spectral densities (PSDs) of target 2-second EEG windows from the BBHI dataset (black) are compared against the PSDs of reconstructed outputs for the LSTMVAE (top, red gradient), VAEEG (middle, blue gradient), and Transformer (bottom, green gradient) models. Three latent space dimensions are shown, $N_{LS} = \{50, 250, 500\}$, with color intensity increasing from lighter to darker shades as $N_{LS}$ increases. Higher $N_{LS}$ improves high-frequency reconstruction fidelity, particularly for the Transformer architecture.
        }
        \label{fig:bbhi_attenuation}
    \end{figure}

    Figure~\ref{fig:bbhi_attenuation} demonstrates that, for the reconstruction autoencoders (LSTMVAE, VAEEG, Transformer), the latent space
    dimensionality acts as a determinant for high-frequency attenuation.
    Smaller latent dimensions impose an information bottleneck that
    disproportionately filters out higher frequencies.
    This poses a significant limitation for tasks requiring high-frequency feature
    stratification; to capture such components, one must ensure a sufficiently
    high-dimensional latent space and ensure accurate high-frequency reconstruction.
    While increasing latent space dimensionality improves high-frequency
    reconstruction, this improvement appears to follow a law of diminishing returns.
    This saturation is consistent with the amplitude-weighted objective implicitly deprioritizing higher frequencies, which typically carry lower power in EEG relative to lower-frequency bands. Consequently, simply expanding the bottleneck may be insufficient to guarantee robust, wide-band feature discovery.
    
    These results provide quantitative and solid grounding for the necessity of alternative strategies, such as specialized loss functions, data augmentation or equalization, multi-scale architectures, etc. to effectively leverage deep learning methodologies in clinical applications where diagnostically relevant information may be concentrated in higher-frequency bands.

\section{Discussion}
    The increasing accessibility of EEG, together with the expansion of open-source
    neuroscience, has facilitated the collection of massive datasets~
    \cite{Obeid2016, Babayan2019, vanDijk2022, Zhang2025}.
    This shift encourages the development of EEG deep learning models
    capable of unsupervised feature extraction for various downstream tasks,
    including classification and generative modeling~\cite{Chen2025}.
    Autoencoders have emerged as a particularly promising framework for this purpose.
    However, a critical question remains: whether raw EEG signals in the temporal
    domain retain sufficient information for nuanced challenges such as the
    stratification of cognitive decline and whether this information can be captured
    by deep-learning models.

    In this study, we addressed the inherent limitations of time-domain autoencoders
    for EEG feature distillation through a series of synthetic experiments,
    cross-referencing these findings with observations from empirical datasets.

    Our synthetic results demonstrate that disentangling frequency-specific
    features is significantly more complex at higher than at lower frequencies.
    Generative factors appear to be more faithfully isolated in lower frequency bands,
    presenting an immediate constraint for autoencoders trained on raw EEG time series.

    While we have corroborated that increasing latent dimensionality ($N_{LS}$)
    enhances high-frequency reconstruction and feature distillation, both in
    synthetic and empirical data, recent research in computer vision suggests that
    excessively large latent spaces may paradoxically degrade the model’s generative
    capabilities~\cite{Xiang2025, Yao2025}, a limitation that needs to be considered
    and adapted according to the downstream task where the embedded features will be
    used.
    Parallel findings in diffusion model research indicate that while decoders
    require high-frequency latent components to recover fine signal details, current
    state-of-the-art encoders often under-represent these contents. This leads to
    insufficient exposure and subsequent underfitting of high-frequency bands
    during training~\cite{Lai2025}, a limitation likely shared by the VAE encoders
    utilized in this work.

    To better understand the origin of this high-frequency under-representation, we next investigated the autoencoders' learning dynamics, visualizing the exact training moments where different frequencies were learned. Our goal was to discern how spectral learning order may interact with the typical decaying spectrum in EEG signals. In the equal-amplitude condition ($A_i=1$), the LSTMVAE exhibited strong spectral ordering, whereas the VAEEG and Transformer architectures captured low and
    high frequencies simultaneously.
    The behavior of the LSTMVAE likely originates from the sequential nature of the
    LSTM layers, where the hidden state serves as a low-pass bottleneck that prioritizes
    long-term dependencies over rapid fluctuations.
    Although convolutional networks can exhibit spectral bias
    \cite{Chakrabarty2019, Fridovic2021}, its absence in the VAEEG likely results
    from the inclusion of residual blocks, which attenuate this effect.
    In the case of the Transformer, the global, distance-agnostic attention
    mechanism allows the network to bypass the sequential processing constraints
    inherent in recurrent models.

    However, under the condition of decaying amplitudes ($A_i \propto 1/f_i^{0.75}$),
    which simulates the spectrum decay in EEG recordings, all three architectures
    showed delayed learning of high-frequency components.
    We argue that this is less a reflection of architectural constraints and more a
    consequence of the loss landscape.
    Because the Mean Squared Error (MSE) loss is sensitive to absolute magnitude,
    components with higher amplitudes (low frequencies) dominate the total error.
    Consequently, gradient descent prioritizes the minimization of these high-power
    components, effectively ``masking`` the architectural capacity of CNNs and
    Transformers to resolve high frequencies early in training.
    Thus, while time-domain feature learning is inherently biased toward slower
    frequencies, learning order is not the sole limiting factor. This is
    evidenced by our finding that the LSTMVAE encoded $\beta / \gamma$ features more
    effectively than the VAEEG or Transformer in our initial cognitive-decline-related slowing synthetic experiment.
    Instead, the primary challenge appears to be the magnitude-driven nature of
    MSE reconstruction loss minimization.

Finally, we extended our analyses to REVE, a state-of-the-art transformer-based masked autoencoder pretrained on more than 60{,}000 hours of EEG recordings. Despite successfully disentangling the generative factors underlying our synthetic datasets, REVE encoded higher-frequency representations less faithfully than lower-frequency ones, as reflected in the reduced accuracy of downstream classifiers on the $\beta/\gamma$ synthetic dataset. Interestingly, REVE showed no evidence of spectral learning order during training, even under the decaying-amplitude condition. By discarding the transformer architecture and masked objective, we hypothesize that this robustness to spectral learning order may be the result of REVE's initial linear patch embedding, which tokenizes each patch into a vector via a single linear transformation, bypassing the sequential or convolutional processing pathways implicated in spectral ordering for the other architectures studied here. However, it is important to note that the absence of training bias does not ensure that all frequencies are equally represented in the resulting latent space. Because EEG power is concentrated in the lower frequencies and the alpha band, unsupervised foundation models may implicitly over-prioritize this dominant content, underrepresenting higher-frequency information that may nonetheless carry clinically relevant signal. This likely accounts for the reduced performance on the $\beta/\gamma$ dataset observed with REVE's frozen pretrained weights.

    Our study counts with limitations that shall be addressed in further
    work to gain deeper insight into the learning dynamics of VAEs for EEG.

    We did not explicitly evaluate the distinction between periodic and
    aperiodic ($1/f$) EEG components~\cite{Donoghue2020, Donoghue2021}.
    While neural networks are capable of capturing novel features, our focus
    remained on periodic oscillations given their established role in reported
    clinical EEG alterations~\cite{Buzski2012, Jeong2004}.

    By focusing on previously validated architectures known for promising EEG
    reconstruction, we prioritized practical relevance and empirical utility over a
    rigorous theoretical analysis of foundational models.
    However, this choice leaves the generative capabilities of these autoencoders
    largely under-explored.
    The integration of diffusion-based models in future research would facilitate a more comprehensive evaluation, particularly concerning how spectral constraints
    impact generative capabilities and the stability of the learned latent space.

    Furthermore, subsequent studies should move toward a quantitative evaluation of
    methodologies designed to specifically improve the reconstruction and feature
    distillation of high-frequency components.
    Several promising strategies have been proposed in recent literature to address
    these imbalances.
    For instance, the inclusion of Mean Squared Error (MSE) between signal
    derivatives within the loss function to better capture rapid temporal
    transients~\cite{Chuang2022}.
    Another approach involves the use of sub-band decomposition, where the signal is
    filtered into discrete bands to train dedicated autoencoders,
    thereby bypassing the dominance of low-frequency power~\cite{Zhao2024} although
    higher frequencies still showed worse reconstruction than slower ones.
    Additionally, ensuring frequency-domain accuracy may be achieved by integrating
    Fourier-based MSE terms directly into the loss
    function~\cite{Bjork2022, Bredell2023}.
    Finally, more methodologically complex alternatives adapted from physics-informed
    neural networks, such as high-frequency scaling, could be applied to ensure that
    high-frequency components contribute equitably to the gradient
    descent process~\cite{Khodakarami2026}.

\section{Conclusions}
In this study, we have systematically investigated the spectral constraints of
unsupervised feature extraction in EEG using deep autoencoders.
Through a combination of synthetic experiments and empirical validation, we have
demonstrated that while autoencoders are highly effective at capturing macro-scale
spectral shifts such as ``alpha slowing``, they remain inherently biased against
higher-frequency components due to the natural $1/f$ EEG spectrum.
Our investigation into learning dynamics confirms that this limitation arises from a generalized magnitude-driven masking effect caused by the Mean Squared Error (MSE) loss function in the presence of $1/f$ decaying spectra.

The results highlight a significant trade-off in the design of EEG foundation
models.
While increasing latent space dimensionality can mitigate spectral attenuation and improve the reconstruction of $\beta$ and $\gamma$ oscillations, this comes at the cost of the compactness and generative stability of the learned latent space. Even so, accurate representational resolution of higher-frequency components is not guaranteed.
Furthermore, we have shown that architectural choices, such as the use of residual connections in CNNs and global attention in Transformers, can partially alleviate spectral learning order, analogous to spectral bias, though they remain constrained by the dominance of high-power low-frequency signals in reconstruction loss minimization.

As the field moves toward large-scale generative models and foundation-level feature extractors, it is imperative to move beyond simple time-domain reconstruction.
Future efforts should prioritize the integration of spectral-domain loss terms and gradient-balancing techniques to ensure that the subtle, high-frequency
transients critical for diagnosing complex brain disorders are not ``smoothed``
out of the latent representation. By addressing these spectral blind spots, deep learning methodologies can better fulfill their promise as robust, objective tools for clinical biomarker discovery.

\section{Acknowledgments}
This work has been partially funded under the project Artificial Intelligence for the Early Diagnosis and Treatment of Highly Prevalent Diseases in Aging project, funded by the R\&D Missions in Artificial Intelligence Program of the Ministry of
496 Economic Affairs and Digital Transformation (MIA.2021.M02.0007). David Aquilué-Llorens is supported by a fellowship from the
”la Caixa” Foundation (ID 100010434). The fellowship code is B006426.

\clearpage

\printbibliography

\clearpage
\section*{Appendix 1: Synthetic data UMAP embeddings for larger Latent Space Dimensions}
    \clearpage

\section*{Appendix 2: Spectral Learning Order Study - Reconstruction evolution}
    \begin{figure}[h!]
        \centering
        \includegraphics[width=\textwidth]{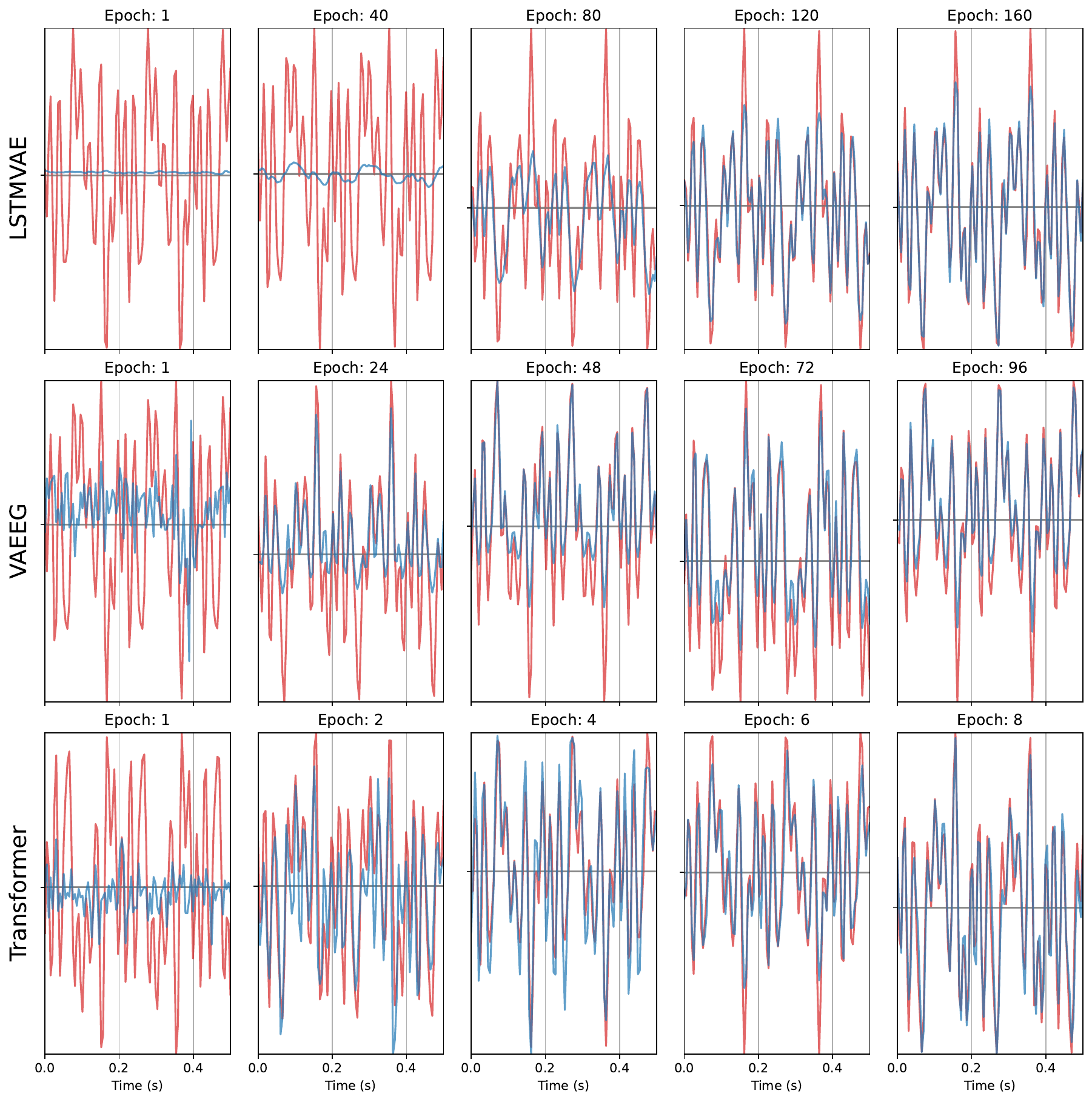}
        \caption{
            \textbf{Spectral Learning Order - Reconstruction Examples}
            Evolution of reconstructed segments in the synthetic experiment adapted
            from Rahaman et al. \cite{Rahaman2018} illustrating the evolution of
            frequency learning as a function of training progress in the
            equal-amplitude condition ($A_i=1$) across three architectures:
            LSTMVAE (upper panels), VAEEG (middle panels), and Transformer (lower
            panels), all with $N_{LS} = 50$
        }
        \label{fig:spec_bias_epochs_50}
    \end{figure}
    \clearpage

\section*{Appendix 3: Latent Structure of Sex, Age, Cognitive Performance}

    \clearpage

\section*{Appendix 4: Spectral Learning Order in  REVE Training}\label{sec:app_reve_spec_bias}

We extended our Rahaman-style~\cite{Rahaman2018} study to the REVE foundation model. This study required training the entire foundation model from scratch, using the synthetic dataset described in section \ref{sec:methods_spectral_bias}. However, due to the significant differences in network architecture and training objectives in the REVE model, a series of modifications were applied to our original Spectral Learning Order study.

\begin{figure}[htbp]
    \centering
    \begin{subfigure}[b]{0.32\textwidth}
        \centering
        \includegraphics[width=\textwidth]{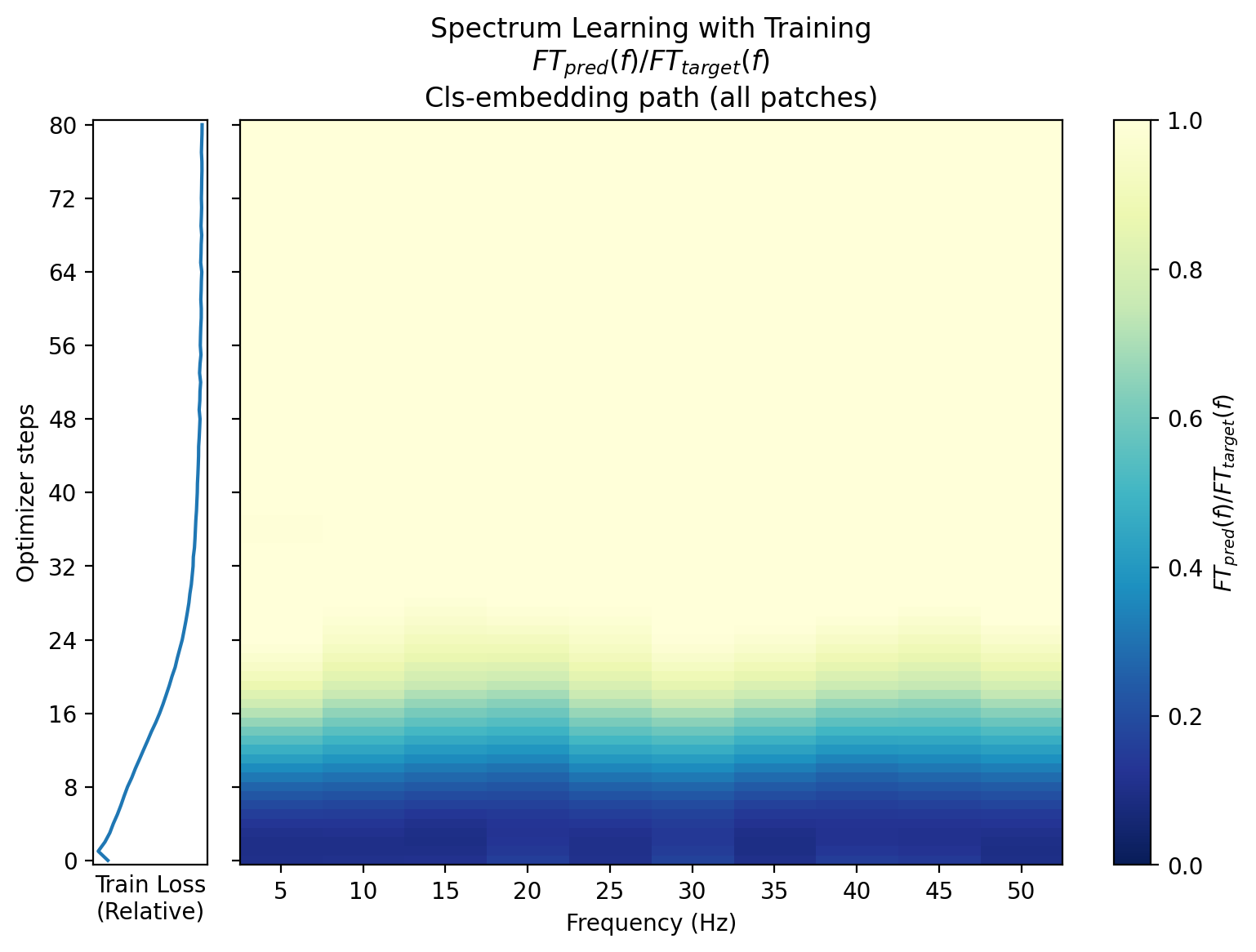}
        \caption{Uniform Amplitude}
        \label{fig:spec_bias_uniform}
    \end{subfigure}
    \hfill
    \begin{subfigure}[b]{0.32\textwidth}
        \centering
        \includegraphics[width=\textwidth]{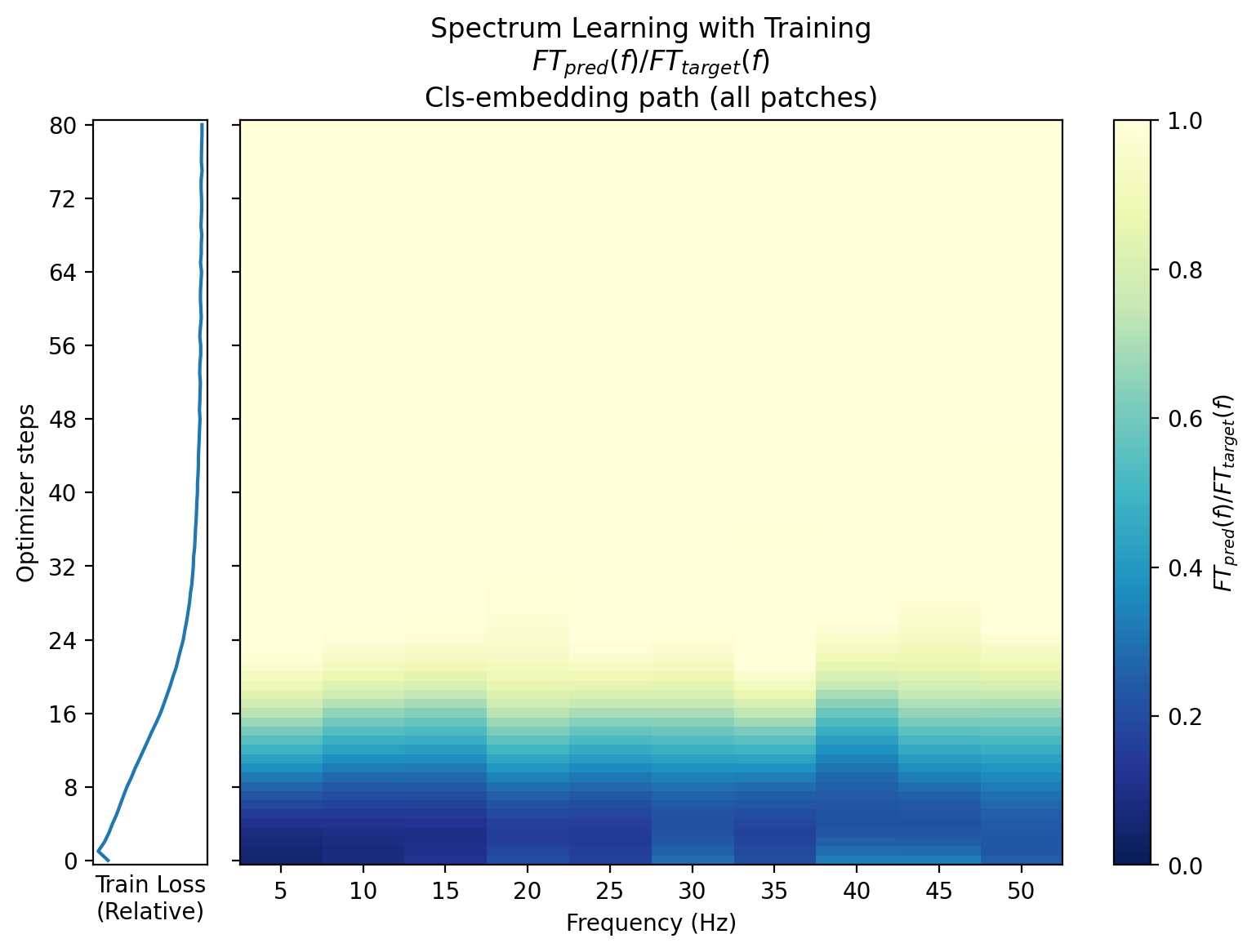}
        \caption{Decaying Amplitude}
        \label{fig:spec_bias_decay}
    \end{subfigure}
    \hfill
    \begin{subfigure}[b]{0.32\textwidth}
        \centering
        \includegraphics[width=\textwidth]{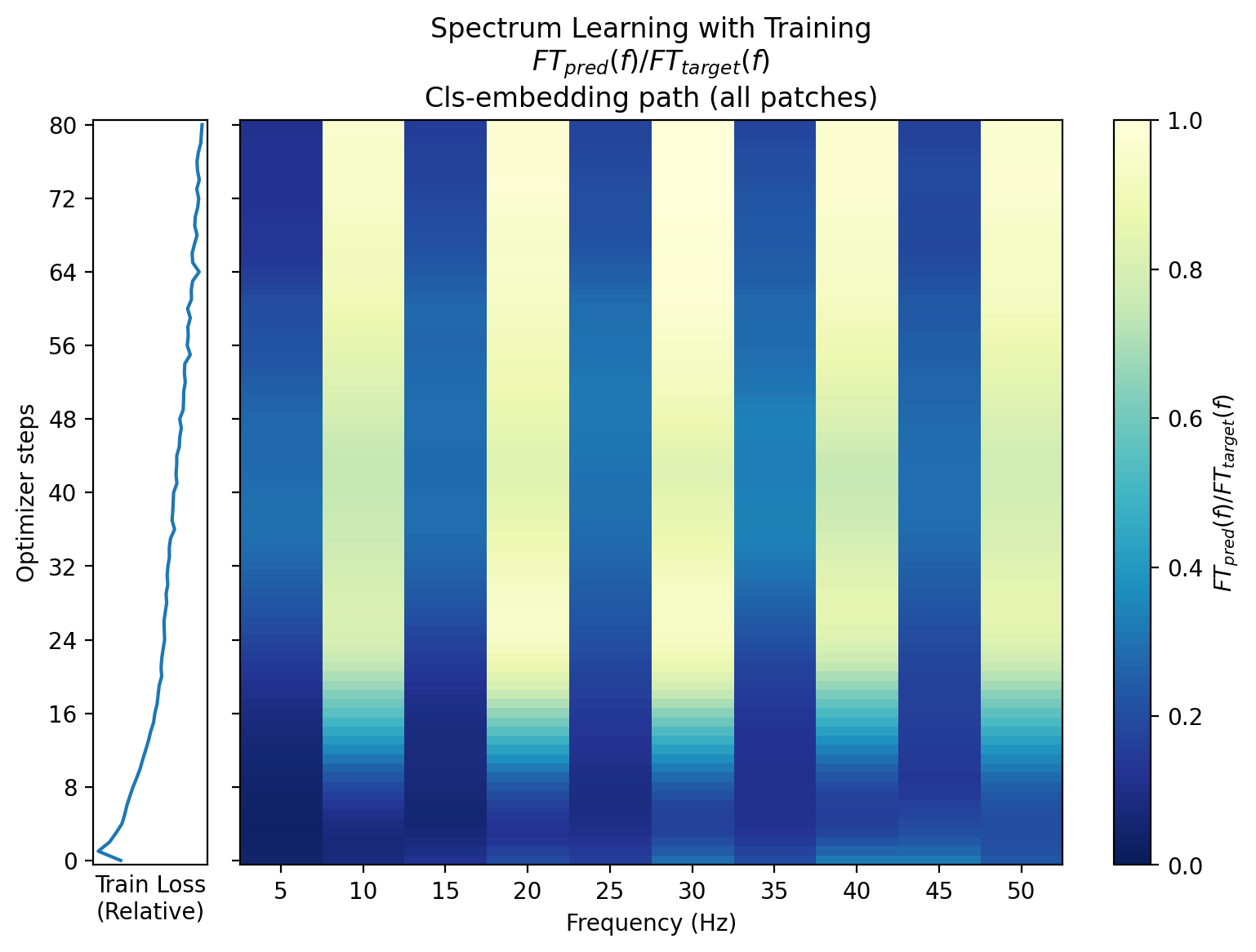}
        \caption{Official patch overlap}
        \label{fig:spec_bias_overlap}
    \end{subfigure}
    \caption{\textbf{Spectral Learning Order Study on REVE}. For Figure details, refer to Figure \ref{fig:spec_bias_50} in the manuscript.}
    \label{fig:spec_bias_REVE}
\end{figure}

We trained REVE using its original paper \textit{"Base"} architecture and objective: a 22-layer, 512-dimensional transformer encoder over 1-second signal patches, a 4-layer decoder, random patch masking (ratio 0.55), and L1 reconstruction loss, optimized with AdamW (with learning rate of $\eta = 5 \times 10^{-5}$, to better elucidate possible spectral ordering effects). We trained the network for 20 epochs and repeated the training with 10 different random initializations. Signals were extended from two to 10 seconds to force dimensionality reduction through a 512-dimensional embedding. Patches were made non-overlapping (stride 1 s instead of the official 0.9 s) because we observed that, with a 0.9 s stride, frequencies that are odd multiples of 5 Hz reverse sign between consecutive patches while even multiples do not, which introduced a parity-reconstruction artifact unrelated to spectral learning order (see Figure\ref{fig:spec_bias_overlap}).

REVE's pretraining objective contains two reconstruction pathways. The primary loss predicts masked patches from the visible tokens through the decoder, which is discarded after pretraining. For the secondary loss, a learned query vector attention-pools the token outputs of every encoder stage into a single 512-dimensional context vector from which a small MLP head, given each masked patch's positional embedding, reconstructs the masked patches. At each optimizer step we computed the spectral fidelity ratio described in Section \ref{sec:methods_spectral_bias} per frequency, averaged over patches and training samples, for both pathways. Since in the autoencoder spectral learning order study the entire signal must pass through a compressed representation in the latent space, we chose the secondary pathway to represent frequency learning during training in Figure \ref{fig:spec_bias_REVE}. Still, some differences are worth noting: the REVE embedding is generally wider (512 vs. 50–250 dimensions, although the $N_{LS}=250$ results seem to indicate small changes in spectral ordering with increased latent size); the secondary pathway is trained jointly with the primary decoder loss at weight $\lambda = 0.1$ rather than as the sole objective; and reconstruction is patch-wise conditioned on a positional code rather than a single decode of the whole signal.

Figure \ref{fig:spec_bias_REVE} shows that the REVE masked autoencoder exhibits no frequency-specific biases during training, even under the decaying-amplitude condition. Given that the Transformer architecture displayed clear signs of spectral learning order in Figure \ref{fig:spec_bias_50}, two hypotheses remain as possible explanations for the observed lack of spectral attenuation in REVE: the masked autoencoding approach itself, and the linear patch embedding of the raw patches. Extending our autoencoder pipeline to a masked autoencoding objective was a natural first step, and we repeated the analysis under a masked patch-prediction scheme. Each input signal was partitioned into non-overlapping patches of fixed length $P=50$ samples, and at every training batch a fresh random subset comprising a fraction $r=0.5$ of these patches was masked by setting the corresponding samples to zero before the signal was passed to the network. Masked patches were resampled independently at each iteration, so that all patches were masked repeatedly over the course of training. The reconstruction term of the loss was evaluated exclusively on the masked patches, while for the variational models the $\beta$-weighted KL term was left unchanged; consistently, the frequency-wise amplitude ratio was accumulated only over the masked region rather than the full signal. All other aspects of training and evaluation—architecture, optimizer, synthetic dataset, and number of runs—remained identical to the unmasked setting.

\begin{figure}[htbp]
    \centering
    \begin{subfigure}[b]{0.4\textwidth}
        \centering
        \includegraphics[width=\textwidth]{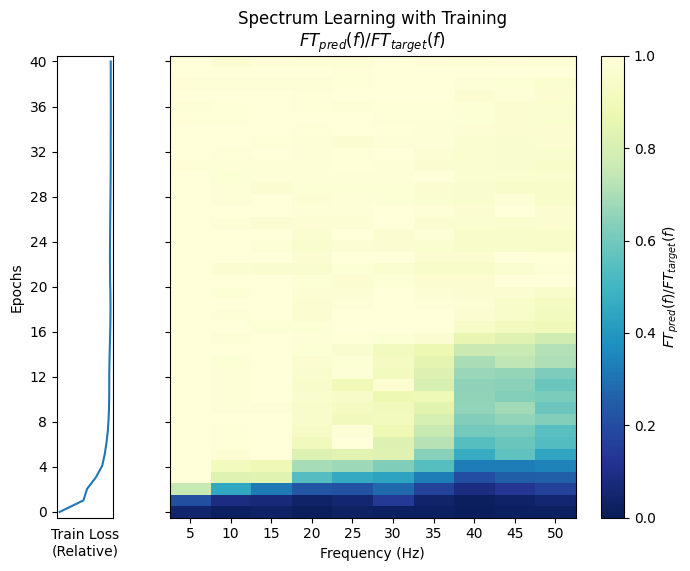}
        \caption{LSTMVAE}
    \end{subfigure}
    \begin{subfigure}[b]{0.4\textwidth}
        \centering
        \includegraphics[width=\textwidth]{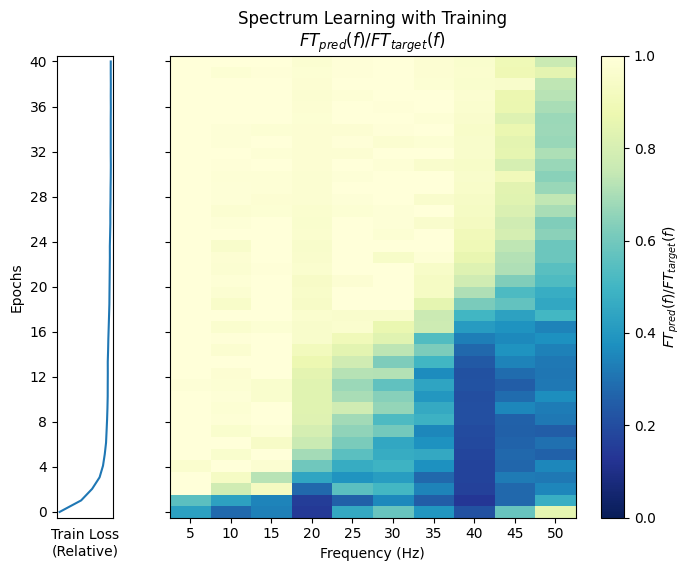}
        \caption{Transformer}
    \end{subfigure}
    \caption{\textbf{Spectral Learning Order Study on LSTMVAE and Transformer reconstruction autoencoders with a modified masked autoencoding objective}. For Figure details, refer to Figure \ref{fig:spec_bias_50} in the manuscript.}
    \label{fig:spec_bias_masked}
\end{figure}

Taken together, these results seem to indicate that the patch linear embedding, used to tokenize raw windows into the embeddings fed to the transformer encoder, is the most likely factor in avoiding frequency-specific biases. Nonetheless, the accuracy achieved by the frozen REVE weights on synthetic data (Figure \ref{fig:synth_data_perf}) indicates that, even in the absence of such biases, higher-frequency features remain harder to disentangle than lower-frequency ones.

\clearpage

\section*{Appendix 5: Topographically-Resolved Spectral Decay for Reconstructing Autoencoders}
    Understanding the spectral fidelity of autoencoder reconstructions is critical for determining whether specific frequency bands are preserved or suppressed during the learning process. The spatial organization of this decay may provide complimentary information of the representational limitations of the models.

    To quantify this behavior, we define a channel-wise Average Spectral
    Attenuation ($\Delta P_{dB}$) metric.
    This measure computes the magnitude of power loss (in dB) within a specified
    frequency range $[f_1, f_2]$, providing a localized assessment of the
    autoencoder’s reconstruction accuracy across the EEG spectrum.

    We define the metric as:
    \begin{equation}\label{eq:attenuation}
        \Delta P_{dB} = \frac{1}{K} \sum_{k \in \text{band}} \left( 10 \log_{10} \frac{P_{out}(f_k)}{P_{in}(f_k)} \right)
    \end{equation}
    where $K$ is the number of frequency bins in the range, and
    $P_{in}(f), P_{out}(f)$ are the averaged (over all epochs and participants in a
    dataset) PSD of the autoencoder's input and reconstruction/output, respectively.

\begin{figure}[htbp]
    \centering
    \begin{subfigure}[b]{0.49\textwidth}
        \centering
        \includegraphics[width=\textwidth]{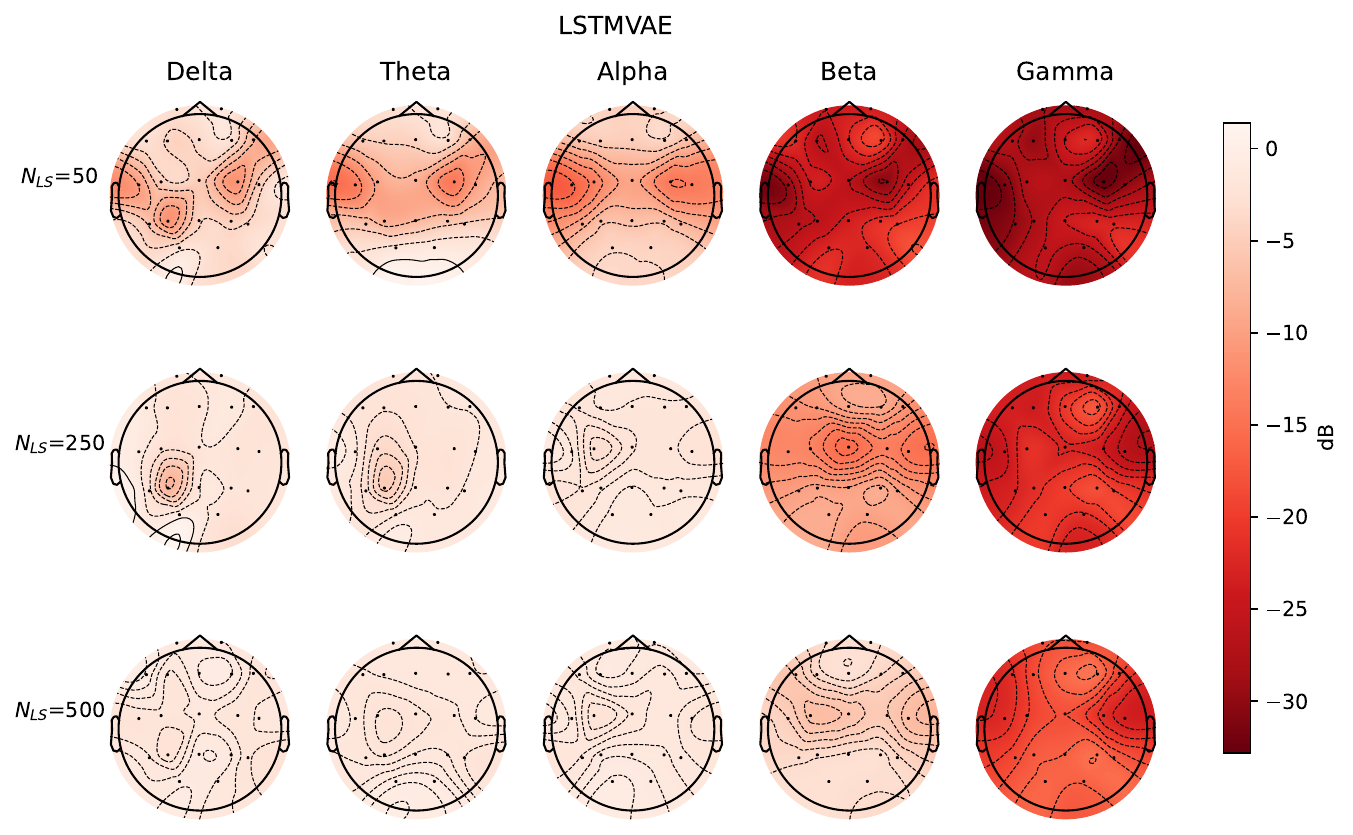}
    \end{subfigure}
    \begin{subfigure}[b]{0.49\textwidth}
        \centering
        \includegraphics[width=\textwidth]{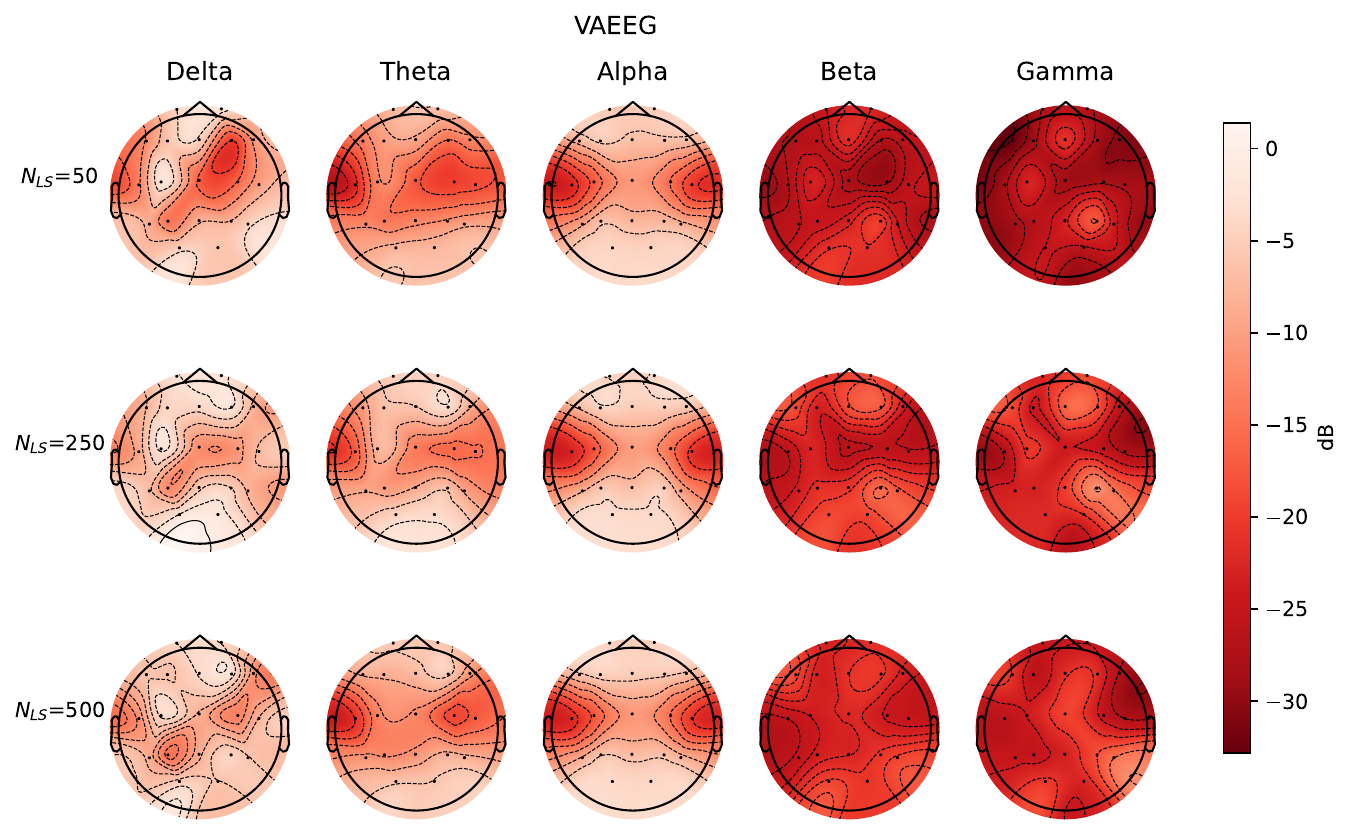}
    \end{subfigure}

    \centering
    \begin{subfigure}[b]{0.49\textwidth}
        \centering
        \includegraphics[width=\textwidth]{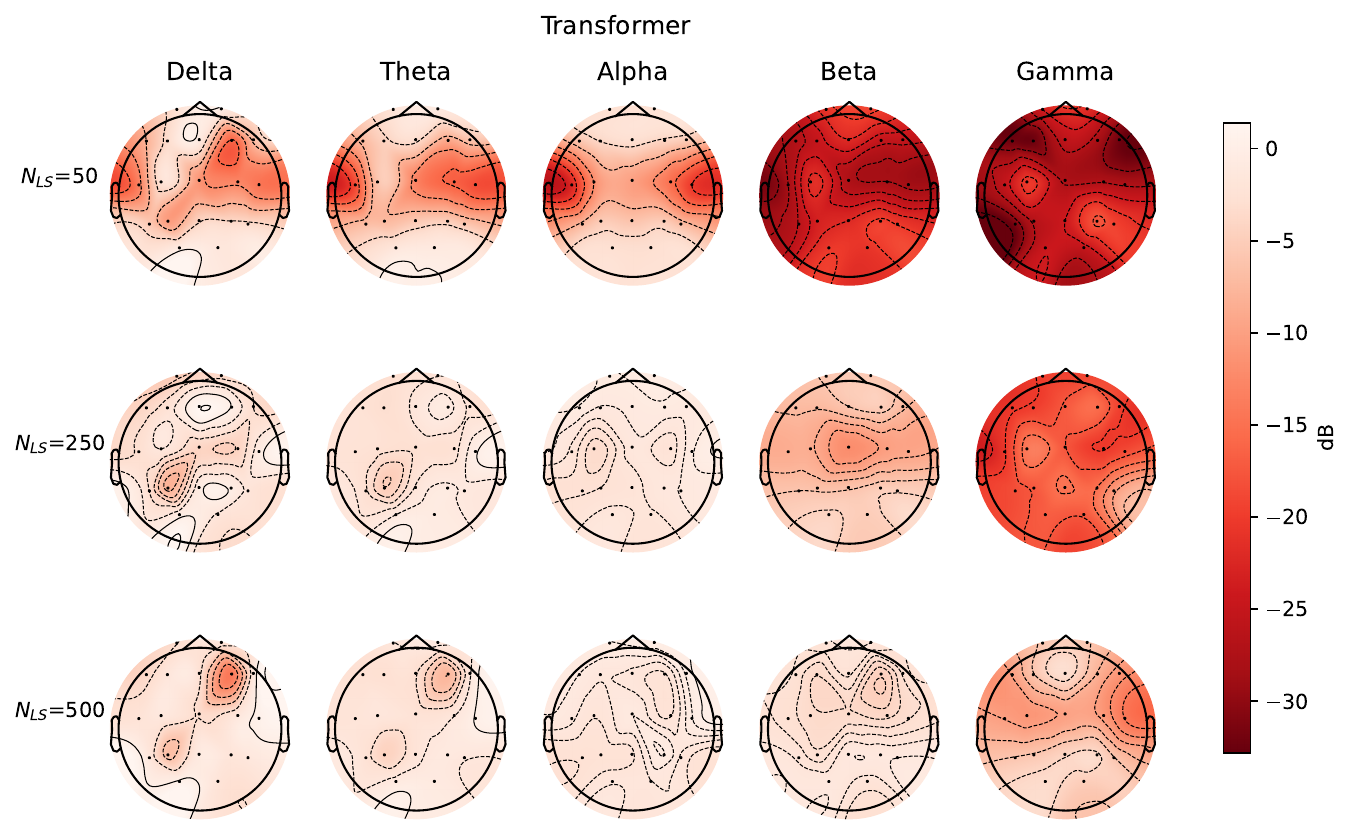}
    \end{subfigure}
    \caption{
    \textbf{Topographical Mapping of Spectral Decay in the held-out test partition of the BBHI Dataset for LSTMVAE (upper left), VAEEG (upper right), and Transformer (lower).}
    Regional distribution of band-specific spectral attenuation
    ($\Delta P_{dB}$) as a function of increasing latent space
    dimensionality ($N_{LS}$). Results are shown for the delta, theta,
    alpha, beta, and gamma bands across the scalp.
    Lighter hues indicate minimized attenuation (higher reconstruction
    fidelity), whereas darker hues denote regions of significant power loss,
    highlighting the spatial and spectral dependencies of the LSTMVAE.
    reconstruction process.}
    \label{fig:app_spec_decay_topo}
\end{figure}

Figure \ref{fig:app_spec_decay_topo} displays the per-channel topographical PSD decay for the three investigated reconstruction autoencoders, resulting from the training and evaluation of the models on the BBHI EEG dataset. Again, increasing Latent Space dimensionality results in attenuating the decay, as it allows for the encoding of higher frequency information. The beta and gamma bands tend to suffer the most from spectral attenuation in the investigated architectures and latent space dimensions. The transformer architecture wiht a $N_{LS} = 500$ achieves the smallest attenuations with satisfactory beta reconstruction.

\end{document}